\documentclass[sigconf,screen]{acmart}
\AtBeginDocument{%
  \providecommand\BibTeX{{%
    \normalfont B\kern-0.5em{\scshape i\kern-0.25em b}\kern-0.8em\TeX}}}

\usepackage{listings}
\usepackage{booktabs}
\usepackage{amsfonts}
\usepackage{graphicx}
\usepackage{hyperref}
\usepackage{textcomp}
\usepackage{listings}
\usepackage{xspace}    
\usepackage{multirow}
\usepackage{multicol}
\usepackage{xcolor} 
\usepackage{amsmath}
\usepackage[linesnumbered,boxed,ruled]{algorithm2e}
\usepackage{listings}
\usepackage{enumitem}

\def \tool{\texttt{DeepPerform}\xspace}

\def \ILFO{\texttt{ILFO}\xspace}

\newcommand{\eg}{{\it e.g.,}\xspace}

\newcommand{\ie}{{\it i.e.,}\xspace}

\newcommand\figref[1]{Fig.~\ref{#1}}

\newcommand\tabref[1]{Table~\ref{#1}}

\newcommand\secref[1]{\S\ref{#1}}
\newcommand\equref[1]{Eq.(\ref{#1})}

\acmConference[ASE '22]{Proceedings of the 37th IEEE/ACM International Conference on Automated Software Engineering}{October 10--14, 2022}{Ann Arbor, Michigan, USA}
\acmYear{2022}
\copyrightyear{2022}
\setcopyright{acmlicensed}
\acmConference[ASE '22]{37th IEEE/ACM International Conference on Automated Software Engineering}{October 10--14, 2022}{Rochester, MI, USA}
\acmBooktitle{37th IEEE/ACM International Conference on Automated Software Engineering (ASE '22), October 10--14, 2022, Rochester, MI, USA}
\acmPrice{15.00}
\acmDOI{10.1145/3551349.3561158}
\acmISBN{978-1-4503-9475-8/22/10}

\ccsdesc[500]{Software and its engineering~Software notations and tools}
\ccsdesc[500]{Computing methodologies~Machine learning}

\keywords{Machine learning, software testing, performance analysis}

\begin{document}

\title{\tool: An Efficient Approach for Performance Testing of Resource-Constrained Neural Networks}

\author{Simin Chen}
\email{simin.chen@UTDallas.edu}
\affiliation{%
  \institution{UT Dallas}
  \city{Dallas}
  \country{USA}
}
\author{Mirazul Haque}
\email{mirazul.haque@utdallas.edu}
\affiliation{%
  \institution{UT Dallas}
  \city{Dallas}
  \country{USA}
}
\author{Cong Liu}
\email{congl@ucr.edu}
\affiliation{%
  \institution{UC Riverside}
  \city{Riverside}
  \country{USA}
}
\author{Wei Yang}
\email{wei.yang@utdallas.edu}
\affiliation{%
  \institution{UT Dallas}
  \city{Dallas}
  \country{USA}
}

\begin{abstract}
Today, an increasing number of Adaptive Deep Neural Networks (AdNNs) are being used on resource-constrained embedded devices.
We observe that, similar to traditional software, redundant computation exists in AdNNs, resulting in considerable performance degradation.
The performance degradation is dependent on the input and is referred to as input-dependent performance bottlenecks (IDPBs).
To ensure an AdNN satisfies the performance requirements of resource-constrained applications, it is essential to conduct performance testing to detect IDPBs in the AdNN. 
Existing neural network testing methods are primarily concerned with correctness testing, which does not involve performance testing.
To fill this gap, we propose \tool, a scalable approach to generate test samples to detect the IDPBs in AdNNs.
We first demonstrate how the problem of generating performance test samples detecting IDPBs can be formulated as an optimization problem.
Following that, we demonstrate how \tool efficiently handles the optimization problem by learning and estimating the distribution of AdNNs' computational consumption.
We evaluate \tool on three widely used datasets against five popular AdNN models. The results show that \tool generates test samples that cause more severe performance degradation~(FLOPs: increase up to 552\%). 
Furthermore, \tool is substantially more efficient than the baseline methods in generating test inputs (runtime overhead: only 6–10 milliseconds).
\end{abstract}

\maketitle

\section{Introduction}
\label{sec:intro}

Deep Neural Networks (DNNs) have shown potential in many applications, such as image classification, image segmentation, and object detection~\cite{ chan2015listen,vaswani2017attention,huang2018yolo}.
However, the power of using DNNs comes at substantial computational costs~\cite{adnn3,adnn4,adnn5,adnn6,adnn7}.
The costs, especially the \textit{inference-time} cost, can be a concern for deploying DNNs on resource-constrained embedded devices such as mobile phones and IoT devices.
To enable deploying DNNs on resource-constrained devices, researchers propose a series of Adaptive Neural Networks~(AdNNs)~\cite{wan2020alert, bateni2018apnet, jiang2021flexible, wang2021asymo, adnn1, adnn2}.
AdNNs selectively activate partial computation units (\eg convolution layer, fully connected layer) for different inputs rather than whole units for computation. The partial unit selection mechanism enables AdNNs to achieve real-time prediction on resource-constrained devices.

Similar to the traditional systems \cite{xiao2013context}, performance bottlenecks also exist in AdNNs. 
Among the performance bottlenecks, some of them can be detected only when given specific input values. Hence, these problems are referred to as input-dependent performance bottlenecks~(IDPBs).
Some IDPBs will cause severe performance degradation and result in catastrophic consequences.
For example, consider an AdNN deployed on a drone for obstacle detection. If AdNNs' energy consumption increases five times suddenly for specific inputs, it will make the drone out of battery in the middle of a trip.
Because of these reasons, conducting performance testing to find IDPB is a crucial step before AdNNs' deployment process.

However, to the best of our knowledge, most of the existing work for testing neural networks are mainly focusing on correctness testing, which can not be applied to performance testing. 
The main difference between correctness testing and performance testing is that correctness testing aims to detect models' incorrect classifications;
while the performance testing is to find IDPBs that trigger performance degradation.
Because incorrect classifications may not lead to performance degradation, existing correctness testing methods can not be applied for performance testing.
To fill this gap and accelerate the process of deploying neural networks on resource-constrained devices, there is a strong need for an automated performance testing framework to find IDPBs.

We identify two main challenges in designing such a performance testing framework. 
First, traditional performance metrics~(\eg latency, energy consumption) are hardware-dependent metrics. Measuring these hardware-dependent metrics requires repeated experiments because of the system noises. Thus, directly applying these hardware-dependent metrics as guidelines to generate test samples would be inefficient. 
Second, AdNNs' performance adjustment strategy is learned from datasets rather than conforming to logic specifications (such as relations between model inputs and outputs). Without a logical relation between AdNNs' inputs and AdNNs' performance, it is challenging to search for inputs that can trigger performance degradation in AdNNs.

To address the above challenges, we propose \tool, which enables efficient performance testing for AdNNs by generating test samples that trigger IDPBs of AdNNs (\tool focuses on the performance testing of latency degradation and energy consumption degradation as these two metrics are critical for performance testing \cite{bateni2020neuos, wan2020alert}).
To address the first challenge, we first conduct a preliminary study~(\secref{sec:study}) to illustrate the relationship between computational complexity~(FLOPs) and hardware-dependent performance metrics~(latency, energy consumption). We then transfer the problem of degrading system performance into increasing AdNNs' computational complexity~(\equref{eq:define}).
To address the second challenge, we apply the a paradigm similar to Generative Adversarial Networks (GANs) to design \tool.
In the training process, \tool learns and approximates the distribution of the samples that require more computational complexity.
After \tool is well trained,  \tool generates test samples that activate more redundant computational units in AdNNs. In addition, because \tool does not require backward propagation during the test sample generation phase, \tool generates test samples much more efficiently, thus more scalable for comprehensive testing on large models and datasets.

To evaluate \tool, we select five widely-used model-dataset pairs as experimental subjects and explore following four perspectives:  \textit{effectiveness}, \textit{efficiency}, \textit{coverage}, and \textit{sensitivity}. 
First, to evaluate the effectiveness of the performance degradation caused by test samples generated by \tool, we measure the increase in computational complexity~(FLOPs) and resource consumption~(latency, energy) caused by the inputs generated by \tool.
For measuring efficiency, we evaluate the online time-overheads  and total time-overheads of \tool in generating different scale samples for different scale experimental subjects. 
For coverage evaluation, we measure the computational units covered by the test inputs generated by \tool.
For sensitivity measurement, we measure how \tool's effectiveness is dependent on the ADNNs' configurations and hardware platforms.
The experimental results show that \tool generated inputs increase  AdNNs' computational FLOPs up to 552\%,  with 6-10 milliseconds overheads for generating one test sample.
We summarize our contribution as follows:

\begin{itemize}

    \item \textbf {Approach. } We propose a learning-based approach~\footnote{\href{https://github.com/SeekingDream/DeepPerform}{https://github.com/SeekingDream/DeepPerform}}, namely \tool, to learn the distribution to generate the test samples for performance testing. Our novel design enables generating test samples more efficiently, thus enable scalable performance testing. 

    \item \textbf {Evaluation.} We evaluate \tool on five AdNN models and three datasets. The evaluation results suggest that \tool finds more severe diverse performance bugs while covering more AdNNs' behaviors, with only 6-10 milliseconds of online overheads for generating test inputs.
    
    \item \textbf{Application.} We demonstrate that developers could benefit from \tool.
    Specifically, developers can use the test samples generated by \tool to train a detector to filter out the inputs requiring high abnormal computational resources~(\S \ref{sec:app}).

\end{itemize}

\section{Background}
\label{sec:background}

\subsection{AdNNs' Working Mechanisms}

\begin{figure}[htp!]%
	\centering
	\includegraphics[width=0.38\textwidth]{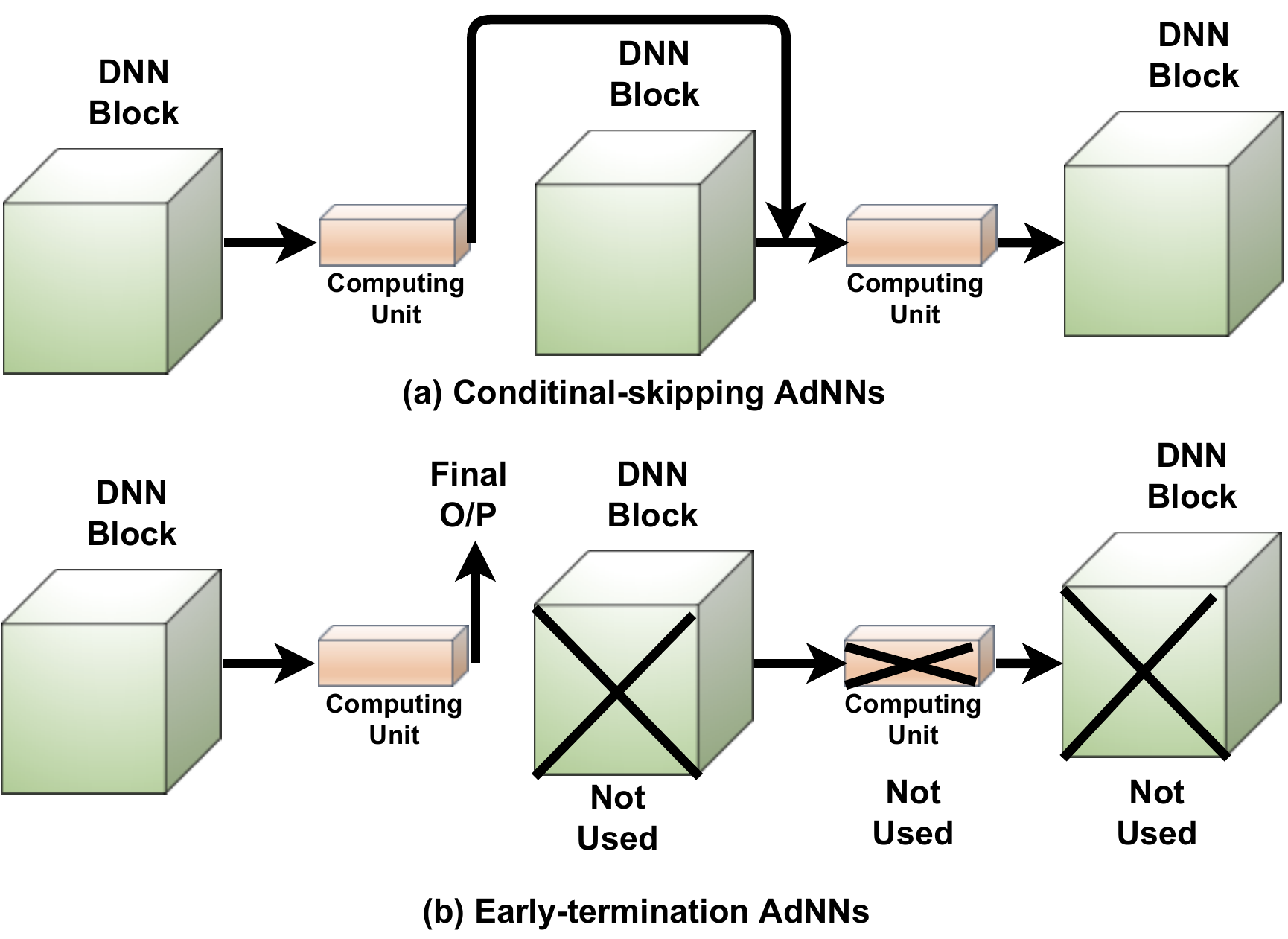}
	\vspace{-3mm}
    \caption{Working mechanism of AdNNs}
	\label{fig:adnn}
\end{figure}

\noindent The main objective of AdNNs~\cite{wan2020alert,teerapittayanon2016branchynet,bolukbasi2017adaptive,kaya2019shallow,wang2018skipnet,adnn1,adnn2,adnn8,adnn9,adnn10} is to balance performance and accuracy. 
As shown in \figref{fig:easy_hard}, AdNNs will allocate more computational resources to inputs with more complex semantics.
AdNNs use intermediate outputs to deactivate specific components of neural networks, thus reducing computing resource consumption.
According to the working mechanism, AdNNs can be divided mainly into two types: \textit{Conditional-skipping AdNNs} and \textit{Early-termination AdNNs}, as shown in \figref{fig:adnn}.
Conditional-skipping AdNNs skip specific layers/blocks if the intermediate outputs provided by specified computing units match predefined criteria. \footnote{a block consists of multiple layers whose output is determined by adding the output of the last layer and input to the block.} (in the case of ResNet).
The working mechanism of the conditional-skipping AdNN can be formulated as:
\begin{equation} \label{eq:new2}
    \begin{cases}
      In_{i+1}=Out_i, \quad \text{if}\ B_i(x) \ge \tau_i \\
      Out_{i+1}=Out_{i}, \quad  \text{otherwise}
    \end{cases}
 \end{equation}
where  $x$ is the input, $In_i$ represents the input of $i^{th}$ layer, $Out_i$ represents the output of $i^{th}$ layer, $B_i$ represents the specified computing unit output of $i^{th}$ layer and $\tau_i$ is the configurable threshold that decides AdNNs' performance-accuracy trade-off mode.
Early-termination AdNNs terminate computation early if the intermediate outputs satisfy a particular criteria.
The working mechanism of early-termination AdNNs can be formulated as,
\begin{equation} \label{eq:new3}
    \begin{cases}
      Exit_{NN}(x)=Exit_i(x), \quad  \text{if}\ B_i(x) \ge \tau_i \\
      In_{i+1}(x)=Out_i(x),\quad\quad   \text{otherwise}
    \end{cases}
 \end{equation}

\subsection{Redundant Computation} 
\label{sec:redundant}
\begin{figure}[ht!]
    \centering
    \includegraphics[width=0.45\textwidth]{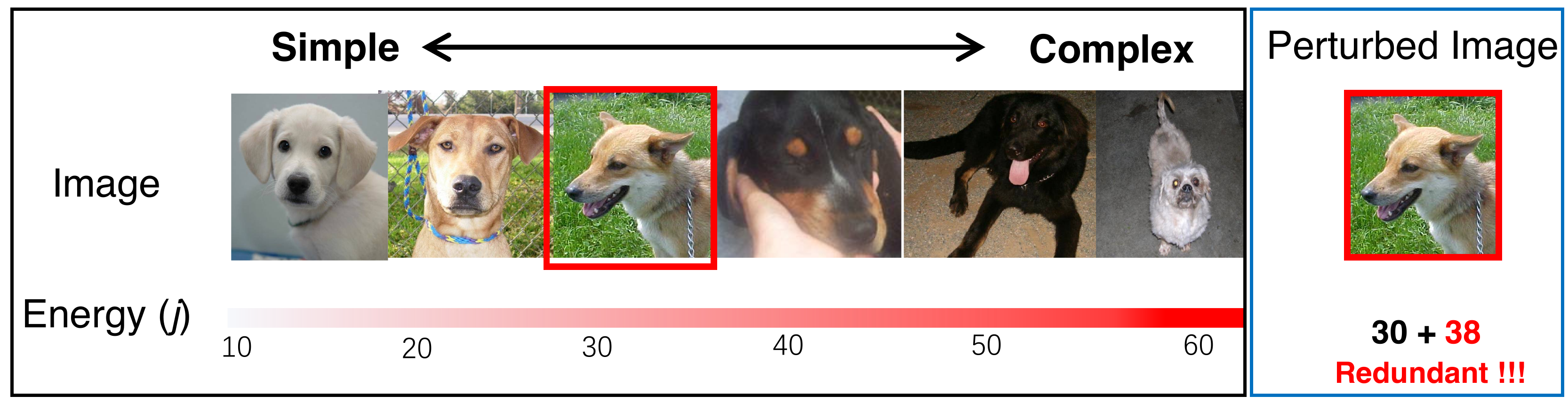}
    \vspace{-3mm}
    \caption{Left Box shows that AdNNs allocate different computational resources for images with different semantic complexity; rights box shows that perturbed image could trigger redundant computation and cause energy surge. }
    \label{fig:easy_hard}
    \vspace{-3mm}
\end{figure}

In a software program, if an operation is not required but performed, we term the operation as redundant operation.
For Adaptive Neural Networks, if a component is activated without affecting AdNNs' final predictions, we define the computation as a redundant computation. 
AdNNs are created based on the philosophy that all the inputs should not require all DNN components for inference.
For example, we can refer to the images in \figref{fig:easy_hard}. 
The left box shows the AdNNs' design philosophy. That is, AdNNs consume more energy for detecting images with further complexity.
 However, when the third image in the left box is perturbed with minimal perturbations and becomes the rightmost one, AdNNs' inference energy consumption will increase significantly (from $30j$ to  $68j$). 
 We refer to such additional computation as redundant computation or performance degradation.

\subsection{Performance \&  Computational Complexity}

In this section, we describe the relationship between hardware-dependent performance metrics and DNN computational complexity.
Although many metrics can reflect DNN performance, we chose latency and energy consumption as hardware-dependent performance metrics because of their critical nature for real-time embedded systems~\cite{bateni2020neuos, wan2020alert}.
Measuring hardware-dependent performance metrics~(\eg latency, energy consumption) usually requires many repeated experiments, which is costly.
Hence, existing work~\cite{wang2018skipnet,adnn1,adnn2,adnn8,adnn9,adnn10} proposes to apply floating point operations~(FLOPs) to represent DNN computational complexity.
However, a recent study \cite{tang2021bridge} demonstrates that simply lowering DNN computational complexity (FLOPs) does not always improve DNN runtime performance.
This is because modern hardware platforms usually apply parallelism to handle DNN floating-point operations (FLOPs).
Parallelism can accelerate computation within layers, while each DNN layer is computed sequentially.
Thus,  For two DNNs with the same total FLOPs, different FLOPs allocating strategies will result in different parallelism utilization and different DNN model performance.
However, for AdNNs, each layer/block usually has a similar structure and FLOPs~\cite{wang2018skipnet, adnn1, adnn2, adnn3}. Thus the parallelism utilization is similar for each block. 
Because parallelism can not accelerate computation between blocks, increasing the number of computational blocks/layers will degrade AdNNs' performance.
To further understand the relation between AdNNs' FLOPs and AdNNs' model performance, we conduct a study in \secref{sec:study}.

\section{Preliminary Study}
\label{sec:study}

\begin{table*}[htbp]
\centering
  \caption{Experiential subject and model performance}
  \vspace{-3mm}
    \label{tab:subject}
    \resizebox{0.95\textwidth}{!}{
    \begin{tabular}{l|l||ccc||ccc|ccc||ccc|ccc}
    \toprule
    \multicolumn{2}{c||}{\multirow{2}[4]{*}{\textbf{Subject} }} & \multicolumn{3}{c||}{\multirow{2}[4]{*}{\textbf{FLOPs}}} & \multicolumn{6}{c||}{CPU  \scriptsize{(Quad-Core ARM® Cortex®-A57 MPCore)}} & \multicolumn{6}{c}{GPU  \scriptsize{(NVIDIA Pascal™ GPU architecture with 256 cores)}} \\ \cline{6-17}   
\multicolumn{2}{c||}{} & \multicolumn{3}{c||}{} & \multicolumn{3}{c|}{Latency} & \multicolumn{3}{c||}{Energy } & \multicolumn{3}{c|}{Latency} & \multicolumn{3}{c}{Energy} \\ 
\cline{1-5}
    \textbf{Dataset}  & \textbf{Model} & Min   & Avg   & Max   & Min   & Avg   & Max   & Min   & Avg   & Max   & Min   & Avg   & Max   & Min   & Avg   & Max \\
    \midrule
    \multirow{2}[4]{*}{\textbf{CIFAR10 (C10)}} & \textbf{SkipNet (SN)} & 195.44  & 248.62  & 336.99  & 0.44  & 0.51  & 0.63  & 65.76  & 76.60  & 316.44  & 0.74  & 0.94  & 1.39  & 168.07  & 245.62  & 439.38  \\
\cmidrule{2-17}          & \textbf{BlockDrop (BD)} & 72.56  & 180.51  & 228.27  & 0.11  & 0.23  & 0.37  & 15.89  & 34.17  & 161.12  & 0.13  & 0.33  & 0.71  & 29.60  & 73.27  & 282.59  \\
    \midrule
    \multirow{2}[2]{*}{\textbf{CIFAR100 (C100)}} & \textbf{DeepShallow (DS)} & 38.68  & 110.47  & 252.22  & 0.04  & 0.11  & 0.25  & 3.47  & 15.32  & 37.81  & 0.09  & 0.37  & 1.08  & 12.63  & 75.49  & 441.60  \\  \cmidrule{2-17} 
          & \textbf{RaNet (RN)} & 31.50  & 41.79  & 188.68  & 0.07  & 0.21  & 2.96  & 8.21  & 27.99  & 448.96  & 0.10  & 0.36  & 5.81  & 15.87  & 60.22  & 997.73  \\
    \midrule
    \textbf{SVHN}  & \textbf{DeepShallow (DS)} & 38.74  & 161.40  & 252.95  & 0.04  & 0.16  & 0.27  & 3.99  & 23.35  & 91.28  & 0.03  & 0.37  & 0.82  & 4.16  & 78.66  & 180.39  \\
    \bottomrule
    \end{tabular}%
    }
  \label{tab:addlabel}%
\end{table*}%

\subsection{Study Approach}
Our intuition is to explore the worst computational complexity of an algorithm or model.
For AdNNs, the basic computation are the floating-point operations~(FLOPs). Thus, we made an assumption that the FLOPs count of an AdNN is a hardware-independent metric to approximate AdNN performance.
To validate such an assumption, we conduct an empirical study.
Specifically, we compute the  \textit{Pearson  Product-moment  Correlation  Co-efficient} (PCCs)~\cite{Rice2006}  between  AdNN FLOPs against  AdNN  latency and energy consumption.
PCCs are widely used in statistical methods to measure the linear correlation between two variables.
PCCs are normalized covariance measurements, ranging from -1 to 1. Higher PCCs indicate that the two variables are more positively related.
If the PCCs between FLOPs against system latency and system energy consumption are both high, then we validate our assumption.

\subsection{Study Model \& Dataset}
We select subjects (\eg model,dataset) following policies below.
\begin{itemize}[leftmargin=*]
    \item The selected subjects are  publicly available.
    \item The selected subjects are widely used in existing work.
    \item The selected dataset and models should be diverse from different perspectives. \eg,
    the selected models should include both early-termination and conditional-skipping AdNNs.
\end{itemize}
We select five popular model-dataset combinations used for image classification tasks as our experimental subjects. The dataset and the corresponding model are listed in \tabref{tab:subject}.
We explain the selected datasets and corresponding models below.

\noindent\textbf{Datasets.} CIFAR-10~\cite{krizhevsky2009learning} is a database for object recognition. There is a total of ten object classes for this dataset, and the image size of the image in CIFAR-10 is $32\times32$.
CIFAR-10 contains 50,000 training images and 10,000 testing images.
CIFAR-100~\cite{krizhevsky2009learning} is similar to CIFAR-10 \cite{krizhevsky2009learning} but with 100 classes. It also contains 50,000 training images and 10,000 testing images.
SVHN \cite{netzer2011reading} is a real-world image dataset obtained from house numbers in Google Street View images.
There are 73257 training images and 26032 testing images in SVHN.

\noindent\textbf{Models.} For CIFAR-10 dataset, we use SkipNet \cite{wang2018skipnet} and BlockDrop \cite{wu2018blockdrop} models. 
SkipNet applies reinforcement learning to train DNNs to skip unnecessary blocks, and BlockDrop trains a policy network to activate partial blocks to save computation costs.
We download trained SkipNet and BlockDrop from the authors' websites.
For CIFAR-100 dataset, we use RaNet \cite{yang2020resolution} and DeepShallow \cite{kaya2019shallow} models for evaluation.
DeepShallow adaptive scales DNN depth, while RaNet scales both input resolution and DNN depth to balance accuracy and performance.
For SVHN dataset, DeepShallow \cite{kaya2019shallow} is used for evaluation.
For RaNet \cite{yang2020resolution} and DeepShallow \cite{kaya2019shallow} architecture, the author does not release the trained model weights but open-source their training codes. Therefore, we follow the authors' instructions to train the model weights.

\subsection{Study Process}

We begin by evaluating each model's computational complexity on the original hold-out test dataset.
After that, we deploy the AdNN model on an Nvidia TX2 \cite{tx2} and measure latency and energy usage.
Through \tabref{tab:subject}, we present the FLOPs, latency, and energy consumption of each AdNN.
We   observe  that  the  model  would cost  a  different  number  of  FLOPs  for  different  test samples, and the variance between each test sample could be significant. For instance, for dataset CIFAR-100 and model RaNet,  the  minimum  FLOPs  are  31.5M, while  the maximum FLOPs are 188.68M. 
\subsection{Study Results}

\begin{table}[tbp!]
\caption{PCCs between FLOPs against latency and energy}
\label{tab:study}
\centering
\resizebox{0.48\textwidth}{!}{
    \begin{tabular}{c|c|ccccc}
    \toprule
    \textbf{Hardware} & \textbf{Metric} & \textbf{SN\_C10} & \textbf{RN\_C100} & \textbf{BD\_C10} & \textbf{DS\_C100} & \textbf{DS\_SVHN} \\
    \midrule
    \multirow{2}[1]{*}{\textbf{CPU}} & \textbf{Latency} & 0.68  & 0.67  & 0.93  & 0.99  & 0.95  \\
          & \textbf{Energy} & 0.65  & 0.64  & 0.93  & 0.98  & 0.95  \\ \midrule
    \multirow{2}[1]{*}{\textbf{GPU}} & \textbf{Latency} & 0.48  & 0.56  & 0.91  & 0.99  & 0.97  \\
          & \textbf{Energy} & 0.53  & 0.64  & 0.91  & 0.99  & 0.97  \\
    \bottomrule
    \end{tabular}%
}
\end{table}

From the PCCs results in \tabref{tab:study}, we have the following observations: (i) The PCCs are more than 0.48 for all subjects. The results imply that FLOPs are positively related to latency and energy consumption in AdNNs~\cite{Rice2006}.
Especially for DS\_C100, the PCC achieves 0.99, which indicates the strong linear relationship between FLOPs and runtime performance.
(ii) The PCCs for the same subject on different hardware devices are remarkably similar (\eg, with an average difference of 0.04). According to the findings, the PCCs between FLOPs and latency/energy consumption are hardware independent.
The statistical observations of PCCs confirm our assumption; that is, the FLOPs of AdNN handling an input is a hardware-independent metric that can approximate AdNN performance on multiple hardware platforms.

\subsection{Motivating Example}
\label{sec:motivation}

\begin{table}[tbp]
  \centering
  \caption{System availability  under performance degradation}
  \vspace{-1mm}
  \resizebox{0.28\textwidth}{!}{
    \begin{tabular}{l|ccc}
    \toprule
    \textbf{Subject} & \textbf{Original} & \textbf{Perturbed} & \textbf{Ratio} \\
    \midrule
    \textbf{SN\_C10} & 10,000 & 6,332 & 0.6332 \\
    \textbf{BD\_C10} & 10,000 & 4,539 & 0.4539 \\
    \textbf{RN\_C100} & 10,000 & 5,232 & 0.5232 \\
    \textbf{DS\_C100} & 10,000 & 3,576 & 0.3576 \\
    \textbf{DS\_SVHN} & 10,000 & 4,145 & 0.4145 \\
    \bottomrule
    \end{tabular}%
  \label{tab:motivation}%
  }
\end{table}%

\noindent
To further understand the necessity of conducting performance testing for AdNNs, we use one real-world example to show the harmful consequences of performance degradation.
In particular, we use \texttt{TorchMobile} to deploy each AdNN model on Samsung Galaxy S9+, an Android device with 6GB RAM and 3500mAh battery capacity.
We randomly select inputs from the original test dataset of each subject (\ie \tabref{tab:subject}) as seed inputs and perturb the selected seed inputs with random perturbation. 
Next, we conduct two experiments (one on the selected seed inputs and another one on the perturbed one) on the phone with the same battery. Specifically, we feed both datasets into AdNN for object classification and record the number of inputs successfully inferred before the battery runs out (We set the initial battery as the battery that can infer 10,000 inputs from the original dataset).
The results are shown in \tabref{tab:motivation}, where the column ``original'' and ``perturbed'' show the number of inputs successfully inferred, and the column ``ratio'' shows the corresponding system availability ratio (\ie the system can successfully complete the percentage of the assigned tasks under performance degradation).
Such experimental results highlight the importance of AdNN performance testing before deployment. Otherwise, AdNNs' performance degradation will endanger the deployed system's availability. 

\section{Approach}

In this section, we introduce the detail design of \tool.

\subsection{Performance Test Samples for AdNNs}
\label{sec:mr}
Following existing work~\cite{MirazILFO, lemieux2018perffuzz, lemieux2018perffuzz}, we define performance test samples as the inputs that require redundant computation and cause performance degradation (\eg higher energy consumption).
Because our work focus on testing AdNNs, we begin by introducing redundant computation in AdNNs.
Like traditional software, existing work~\cite{kaya2019shallow,MirazILFO} has shown redundant computation also exist in AdNNs.
Formally, let $g_{f}(\cdot)$ denotes the function that measures the computational complexity of neural network $f(\cdot)$, and  $T_I(\cdot)$ denotes a semantic-equivalent transformation in the input domain.
As the example in \figref{fig:easy_hard}, $T_I(\cdot)$ could be changing some unnoticeable pixels in the input images.
If $g_{f}(T_I(x_i)) > g_{f}(x_i)$ and $f(x_i)$ is correctly computed, then there exist redundant computation in the model $f(\cdot)$ handling $T_I(x_i)$.
In this paper, we consider unnoticeable perturbations as our transformations $T_I(\cdot)$, the same as the existing work~\cite{carlini2017towards, jia2017adversarial, MirazILFO}.
Finally, we formulate our objective to generate performance test samples as searching such unnoticeable input transformation $T_I(\cdot)$, as shown in \equref{eq:define}.
\begin{equation}
    \label{eq:define}
    \begin{split}
        &  \qquad g(T_I(x)) >> g(x)  \\
        & T_I(x) = \{x + \delta(x)  | \quad ||\delta(x)||_p \le \epsilon \}
    \end{split}
\end{equation}

\subsection{\tool Framework}

\begin{figure}[htbp!]
    \centering
    \includegraphics[width=0.45\textwidth]{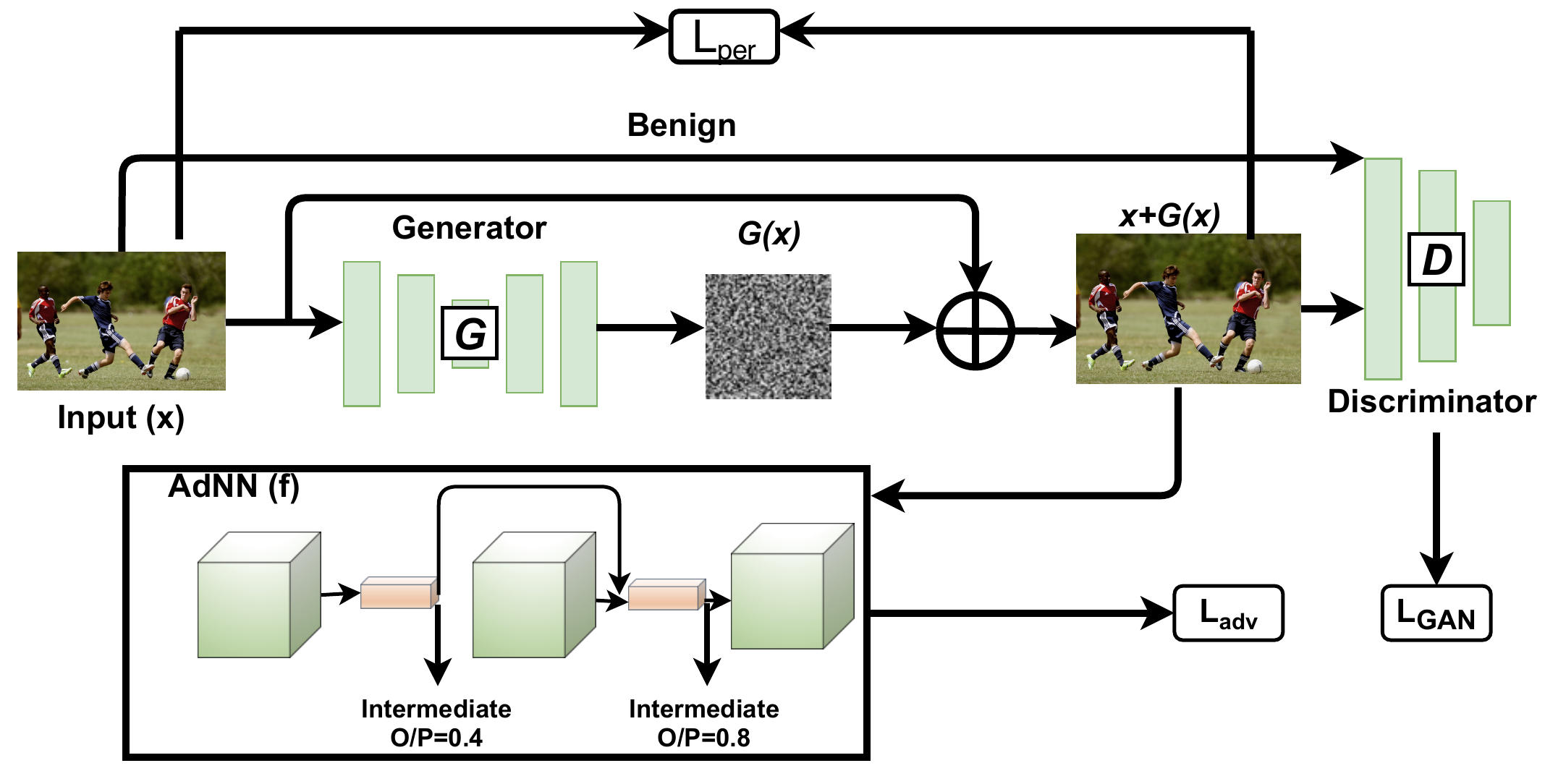}
    \vspace{-1mm}
    \caption{Design overview of \tool}
    \label{fig:overview}
    \vspace{-3mm}
\end{figure}

\figref{fig:overview} illustrates the overall architecture of \tool, which is based on the paradigm of Generative Adversarial Networks (GANs). GANs mainly consist of a generator $\mathcal{G}(\cdot)$ and a discriminator $\mathcal{D}(\cdot)$.
The input $x$ of the generator $\mathcal{G}(\cdot)$ is a seed input and the output $\mathcal{G}(x)$ is a minimal perturbation (\ie $\delta(x)$ in \equref{eq:define}).
After applying the generated perturbation to the seed input, the test sample $T_{I}(x)= x+\mathcal{G}(x)$ is sent to the discriminator.
The discriminator $\mathcal{D}(\cdot)$ is designed to distinguish the generated test samples $x+\mathcal{G}(x)$ and the original samples $x$. 
After training, the generator would generate more unnoticeable perturbation, correspondingly, the discriminator would also be more accurate in distinguishing original samples and generated samples. 
After being well trained, the discriminator and the generator would reach a Nash Equilibrium, which implies the generated test samples are challenging to be distinguished from the original samples.
\begin{equation}
\label{eq:gan_loss}
    \mathcal{L}_{GAN} = \mathbb{E}_{x} log\mathcal{D}(x) + \mathbb{E}_{x} log[1 - \mathcal{D}(x + \mathcal{G}(x))]
\end{equation}
The loss function of the Generative Adversarial Networks~(GANs) can be formulated as Equation \ref{eq:gan_loss}.
In Equation \ref{eq:gan_loss}, the discriminator $\mathcal{D}$ tries to distinguish the generated samples $ \mathcal{G}(x) + x$ and the original sample $x$, so as to encourage the samples generated by $\mathcal{G}$ close to the distribution of the original sample. 

However, the perturbation generated by $\mathcal{G}$ may not be able to trigger performance degradation. To fulfil that purpose, we add target AdNN $f(\cdot)$ into the \tool architecture. While training $\mathcal{G}(\cdot)$, the generated input is fed to AdNN to create an objective function that will help increase the AdNNs' FLOPs consumption.
To generate perturbation that triggers performance degradation in AdNNs, we incorporate two more loss functions other than $\mathcal{L}_{GAN}$ for training  $\mathcal{G}(\cdot)$. 
As  shown  in  \equref{eq:define},  to increase the redundant computation, the first step is to model the function $g_f(\cdot)$.
According to our statistical results in \S \ref{sec:study}, FLOPs could be applied as a hardware-independent metric to approximate AdNNs system performance.
Then we model $g_f(\cdot)$ as  \equref{eq:performance_func}.
\begin{equation}
\label{eq:performance_func}
    g_f(x) = \sum_{i=1}^N W_i \times \mathbb I( B_i(x) > \tau_i)
\end{equation}
Where $W_i$ is the FLOPs in the $i^{th}$ block, $B_i(x)$ is the probability that the $i^{th}$ block is activated,  $\mathbb I(\cdot)$ is the indicator function,  and $\tau_i$ is the pre-set threshold based on available computational resources.
\begin{equation}
\label{eq:adv_loss}
    \mathcal{L}_{adv} =   \ell(g_{f}(x), \; \sum_{i=1}^N W_i)
\end{equation}
To enforce $\mathcal{G}$ could generate perturbation that trigger IDPB, we define our performance degradation objective function as Equation \ref{eq:adv_loss}. 
Where $\ell$ is the Mean Squared Error.
Recall $\sum_{i=1}^N W_i$ is the status that all blocks are activated, then $\mathcal{L}_{adv}$ would encourage the perturbed input to activate all blocks of the model, thus triggering IDPBs.
\begin{equation}
\label{eq:per_loss}
    \mathcal{L}_{per} = \mathbb{E}_x || \mathcal{G}(x) ||_p
\end{equation}

To bound the magnitude of the perturbation, we follow the existing work~\cite{carlini2017towards} to add a loss of the $L_p$ norm of the semantic-equivalent perturbation. Finally, our full objective can be denoted as
\begin{equation}
    \mathcal{L} = \mathcal{L}_{GAN} + \alpha  \mathcal{L}_{adv} + \beta \mathcal{L}_{per}
    \label{eq:obj}
\end{equation}

Where $\alpha$ and $\beta$ are two hyper-parameters that balance the importance of each objective. 
Notice that the goal of the correctness-based testing methods' objective function is to maximize the errors while our objective function is to maximize the computational complexity. Thus, our objective function in \equref{eq:obj} can not be replaced by the objective function proposed in correctness-based testing \cite{carlini2017towards, pei2017deepxplore, tian2018deeptest}.

\subsection{Architecture Details}

\begin{figure}[tbp!]
    \centering
    \includegraphics[width=0.42\textwidth]{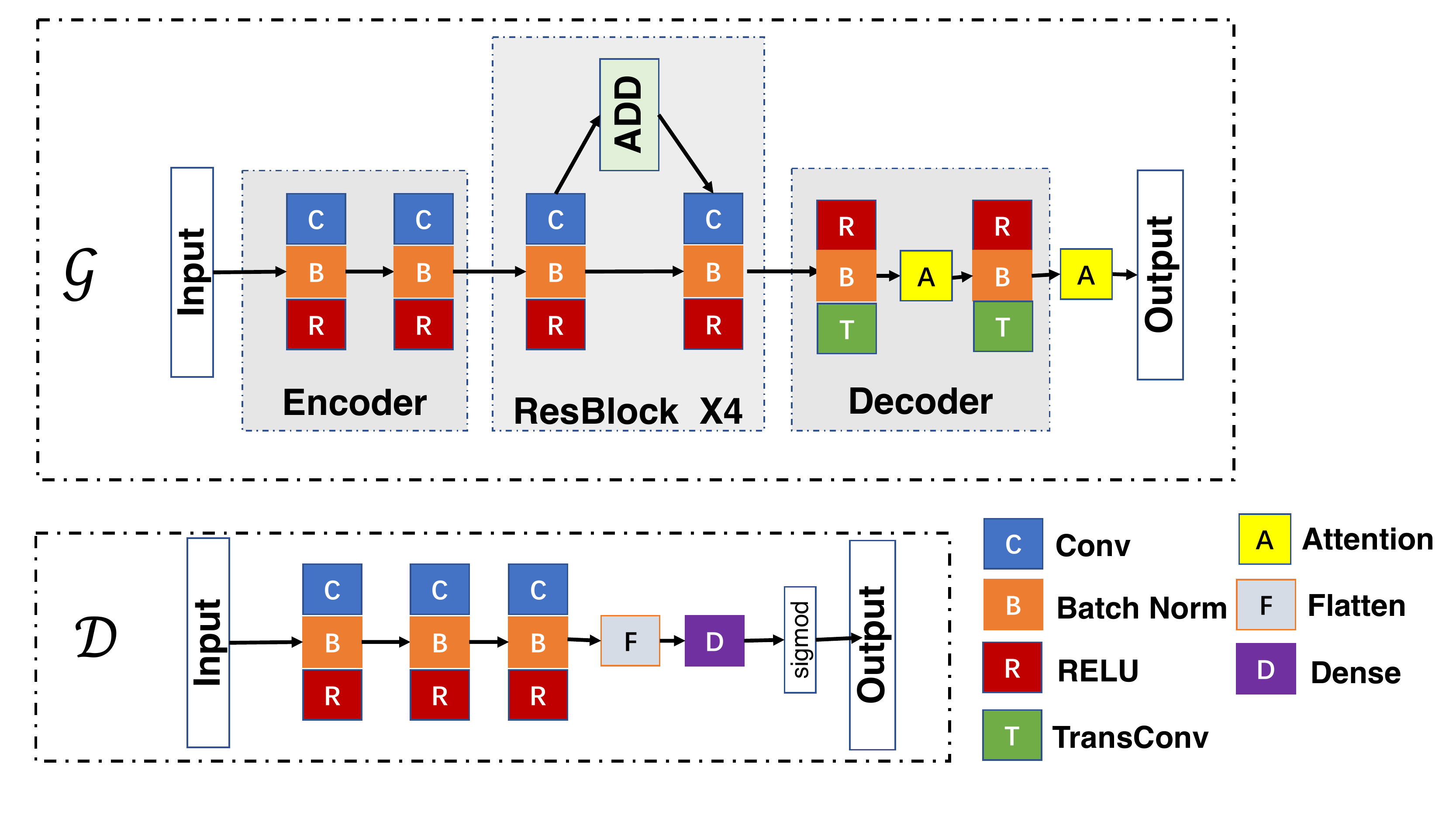}
    \caption{Architecture of the generator and  discriminator }
    \label{fig:arch}
    \vspace{-1mm}
\end{figure}

In this section, we introduce the detailed architecture of the generator and the discriminator.
Our generator $\mathcal{G}$ adapts the structure of encoder-decoder, and the architecture of the discriminator is a  convolutional neural network.
The architectures of the generator and the discriminator are displayed in \figref{fig:arch}.

\noindent\textbf{Generator.}
As shown in \figref{fig:arch}, there are three main components in the generator, that is, the Encoder, the ResBlocks, and the Decoder.
The Encoder repeats the convolutional blocks twice, a convolutional block includes a convolutional layer, a batch normalization layer, and a RELU activation layer.
After the encoding process, the input would be smaller in size but with deep channels.
The ResBlock stacks four residual  blocks~\cite{han2015deep}, which is widely used to avoid the gradient vanishing problem.
The Decoder is the reverse process of the Encoder, the transpose convolutional layer is corresponding to the convolutional layer in the Encoder.
After the decoding process, the intermediate values will be changed back to the same size as the original input to ensure the generated perturbation to be applied to the original seed input.

\noindent\textbf{Discriminator.} The architecture of the discriminator is simpler than the generator. 
There are three convolutional blocks to extract the feature of the input, after that, following a flatten layer and a dense layer for classification.

\begin{algorithm}[tbp]

\LinesNumbered
\scriptsize
  \caption{Training \tool}\label{alg:L2}
\KwIn{The subject AdNNs $f(\cdot)$ to be tested}
\KwIn{Perturbation Constraints $\epsilon$, Perturbation norm $p$  }
\KwIn{Training dataset  $\mathcal{X}$}
\KwIn{Hyper-parameters  $\alpha$, $\beta$}
\KwIn{Maximum  training epochs $T$}
\KwOut{Generator $\mathcal{G}$ and Discriminator $\mathcal{D}$}
$g_f(\cdot)$ = ModelPerformance($f$) \tcp{Construct $g_f$ through Equation \ref{eq:performance_func}.}
\For{epoch in range(0, T)} {
    \For {batch in $\mathcal{X}$}{
     $\overline{x}$ = $\mathcal{G}(x) + x$  \tcp*{generate  test samples}
     $\overline{x}$ = CLIP($\overline{x}, x, p, \epsilon$)  \tcp*{clip test samples}
   $\mathcal{L}_{GAN} += $ ComputeGanLoss($\overline{x}$, $x$, $\mathcal{D}$)  \tcp*{Equation \ref{eq:gan_loss}}
   $\mathcal{L}_{per} += $ ComputePerLoss($\overline{x}$, $x$)  \tcp*{Equation \ref{eq:per_loss}}
   $\mathcal{L}_{adv} += $ ComputeAdvLoss($\overline{x}$, $x$)  \tcp*{Equation \ref{eq:adv_loss}}
    }
    $\triangledown \mathcal{G}$ = ComputeGrad($\mathcal{L}_{GAN} + \mathcal{L}_{per} + \mathcal{L}_{adv}$) \tcp*{Compute  $\mathcal{G}$ gradient}
    $\triangledown \mathcal{D}$ =  ComputeGrad($\mathcal{L}_{GAN}$)\tcp*{Compute $\mathcal{D}$  gradient}
    $\mathcal{G} = \mathcal{G} + \triangledown \mathcal{G}$, $\mathcal{D} = \mathcal{D} + \triangledown \mathcal{D}$\tcp*{Update the weights of $\mathcal{D}$ and $\mathcal{G}$}
}
\end{algorithm}

\subsection{Training \tool}

The training of \tool is comprised of two parts: training the discriminator $\mathcal{D}$ and training the generator $\mathcal{G}$.
Algorithm \ref{alg:L2} explains the training procedure of the \tool. 
The inputs of our algorithm include the target AdNNs $f(\cdot)$, perturbation constraints $\epsilon$, training dataset $\mathcal{X}$, hyper-parameters  $\alpha, \beta$ and max epochs $T$. 
The outputs of our training algorithm include a well-trained generator and discriminator.
First, the algorithm constructs the performance function $g(\cdot)$ through Equation \ref{eq:performance_func} (Line 1).
Then we run $T$ epochs. For each epoch, we iteratively select small batches from the training dataset~(Line 2, 3).
For each seed $x$ in the selected batches, we generate test sample $\overline{x}$ and compute the corresponding loss through \equref{eq:gan_loss}, \eqref{eq:per_loss}, \eqref{eq:adv_loss} (Line 6-8).
We compute the gradients of $\mathcal{G}$ and $\mathcal{D}$ with the computed loss~(Line 10, 11), then we update the weights of $\mathcal{G}$ and $\mathcal{D}$ with the gradients~(Line 12).
The update process is performed iteratively until the maximum epoch is reached.

\section{Evaluation}
\label{sec:evaluation}

We evaluate \tool and answer the following questions:

\begin{itemize}[leftmargin=*]
\item \textit{\textbf{RQ1 (Efficiency)}}: How efficiently does \tool generate test samples? 
    \item \textit{\textbf{RQ2 (Effectiveness)}}: How effective can \tool generate test samples that degraded AdNNs' performance?
    \item \textit{\textbf{RQ3 (Coverage)}}: Can \tool generate test samples that cover AdNNs' more computational behaviors?

    \item \textit{\textbf{RQ4 (Sensitivity)}}: Can \tool behave stably under different settings? 

    \item \textit{\textbf{RQ5 (Quality)}}: What is the semantic quality of the generated test inputs, and how does it relate to performance degradation?

 \end{itemize}

\subsection{Experimental Setup}
\label{sec:setup}

\subsubsection{Experimental Subjects}
We select the five subjects used in our preliminary study~(\secref{sec:study}) as our experimental subjects. 
As we discussed in \secref{sec:study}, the selected subjects are widely used, open-source, and diverse in working mechanisms.

\subsubsection{Comparison Baselines} As we mentioned in \S \ref{sec:background}, almost all existing DNN testing work focuses on correctness testing. As far as we know, \ILFO \cite{MirazILFO} is the state-of-the-art approach for generating inputs to increase AdNNs computational complexity. Furthermore, \ILFO has proved that its backward-propagation approach is more effective and efficient than the traditional symbolic execution (\ie SMT); thus, we compare our method to \ILFO.
\ILFO iteratively applies the backward propagation to perturb seed inputs to generate test inputs. 
However, the high overheads of iterations make \ILFO a time-consuming approach for generating test samples.
Instead of iterative backward computation, \tool learns the AdNNs' computational complexity in the training step. After \tool is trained, \tool applies forward propagation once to generate one test sample.

\subsubsection{Experiment Process} 
\label{sec:process}
We conduct an experiment on the selected five subjects,  
and we use the results to answer all five RQs.
The experimental process can be divided into test sample generation and performance testing procedures.

\noindent\textbf{Test Sample Generation}. For each experimental subject, we split train/test datasets according to the standard procedure\cite{krizhevsky2009learning, netzer2011reading}.
Next, we train \tool with the corresponding training datasets. The training is conducted on a Linux server with three Intel Xeon E5-2660 v3 CPUs @2.60GHz, eight 1080Ti Nvidia GPUs, and 500GB RAM, running Ubuntu 14.04. We  configure the training process with  100  maximum epochs,  0.0001  learning  rate,  and  apply early-stopping techniques~\cite{Yao2007}. We set the hyper-parameter $\alpha$ and $\beta$ as $1$ and $0.001$, as we observe $\mathcal{L}_{per}$ is about three magnitude larger than $\mathcal{L}_{adv}$.
After \tool is trained, we randomly select 1,000 inputs from original test dataset as seed inputs. Then, we feed the seed inputs into \tool to generate test inputs~($x + \mathcal{G}(x)$ in \figref{fig:overview}) to trigger AdNNs' performance degradation.
In our experiments, we consider both $L_2$ and $L_{inf}$ perturbations \cite{carlini2017towards} and train two version of \tool for input generation.
After \tool is trained, we apply the clip operation \cite{li2018adversarial} on $x + \mathcal{G}(x)$ to ensure the generated test sample satisfy the semantic constraints in \equref{eq:define}.

\noindent\textbf{Performance Testing Procedure}. 
For the testing procedure, we select Nvidia Jetson TX2 as our main hardware platform~(We evaluate \tool on different  hardwares in \secref{sec:sensitive}). 
Nvidia Jetson TX2 is a popular and widely-used hardware platform for edge computing, which is built around an Nvidia Pascal-family GPU and loaded with 8GB of memory and 59.7GB/s of memory bandwidth.
We first deploy the AdNNs on Nvidia Jetson TX2.
Next, we feed the generated test samples~(from \tool and baseline) to AdNNs, and measure the response latency and energy consumption~(energy is measured through Nvidia power monitoring tool). 
Finally, we run AdNNs at least ten times to infer each generated test sample to ensure the results are accurate.

\noindent\textbf{RQ Specific Configuration.} 
For RQ1, 2 and 3, we follow existing work~\cite{madry2017towards, MirazILFO, athalye2018obfuscated} and set the maximum perturbations as 10 and 0.03 for $L_2$ and $L_{inf}$ norm separately for our approach and baselines. 
We then conduct experiments in \S \ref{sec:perturbation} to study how different maximum perturbations would affect the performance degradation.
\ILFO needs to configure maximum iteration number and balance weight, we set the maximum iteration number as 300 and the balance weight as $10^{-6}$, as suggested by the authors~\cite{MirazILFO}.
As we discussed in \S \ref{sec:background}, AdNNs require a configurable parameter/threshold to decide the working mode. Different working modes have different tradeoffs between accuracy and computation costs.
In our deployment experiments~(RQ2), we follow the authors~\cite{MirazILFO} to set the threshold as 0.5 for all the experimental AdNNs, and we evaluate how different threshold will affect \tool effectiveness in \secref{sec:sensitive}.
Besides that, to ensure the available computational resources are the same, we run only the AdNNs application in the system during our performance testing procedure.

\subsection{Efficiency}
\label{sec:efficiency}

In this section, we evaluate the efficiency of \tool in generating test samples compared with selected baselines.

\noindent\textbf{Metrics.} We record the \textit{online time overheads} of the test sample generation process (overheads of running $\mathcal{G}$ to generate perturbation), and use the mean online time overhead~(s) as our evaluation metrics. 
A lower time overhead implies that it is more efficient, thus better in generating large-scale test samples.
Because \tool requires training the generator $\mathcal{G}(\cdot)$, for a fair comparison, we also evaluate the \textit{total time overheads}~($\mathcal{G}(\cdot)$ training + test samples generation) of generating different scale numbers of test inputs.

\begin{figure}[htbp!]
    \centering
    \includegraphics[width=0.49\textwidth]{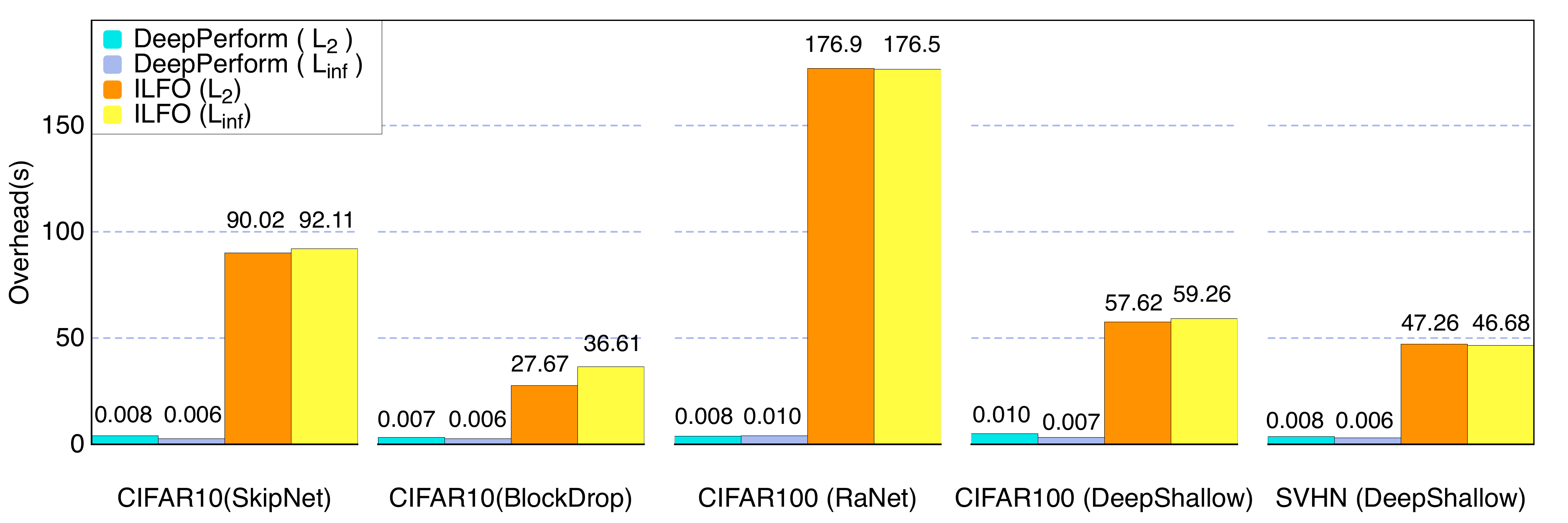}
    \vspace{-6mm}
    \caption{Online overheads to generate one test sample (s)}
    \label{fig:overhead}
\end{figure}

\noindent\textbf{Online Overheads.} The average time overheads of generating one test sample are shown in  \figref{fig:overhead}.
The results show that \tool costs less than 0.01s to generate a test sample under all experimental settings.
In contrast, \texttt{ILFO} requires 27.67-176.9s to generate one test sample. 
The findings suggest that given same time budget, \tool can generate 3952-22112$\times$ more inputs than existing method.
Another interesting observation is that the overheads of  \texttt{ILFO} fluctuate among different subjects, but the overheads of \tool remain relatively constant.
The reason is that the overheads of \tool mainly come from the inference process of the generator, while the overheads of \texttt{ILFO} mainly come from backward propagation.
Because backward propagation overheads are proportional to model size (\ie, a larger model demands more backward propagation overheads), the results of \texttt{ILFO} show a significant variation. 
The overhead of \tool is stable, as its overheads have no relation to the AdNN model size.
The result suggests that when testing large models, \texttt{ILFO} will run into scalability issues, whereas \tool will not.

\begin{figure*}
    \centering
    \includegraphics[width=0.88\textwidth]{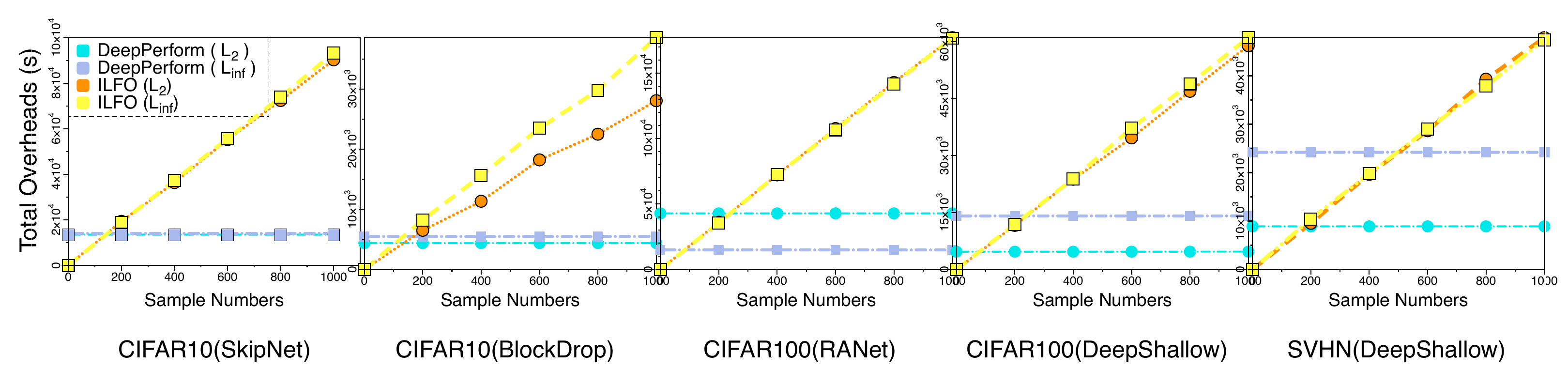}
    \caption{Total overheads of generating different scale test samples (s)}
    \label{fig:total_overhead}
\end{figure*}

\noindent\textbf{Total Overheads.} The total time overheads of generating various scale test samples are shown in \figref{fig:total_overhead}.
We can see from the results that \texttt{ILFO} is more efficient than \tool when the number of generated test samples is minimal (less than 200).
However, when the number of generated test samples grows greater, the overall time overheads of \tool are significantly lower than \texttt{ILFO}.
To create 1000 test samples for SN\_C10, for example, \texttt{ILFO} will cost five times the overall overheads of \tool.
Because the overhead of \texttt{ILFO} is determined by the number of generated test samples \cite{MirazILFO}, the total overheads quickly rise as the number of generated test samples rises.
The main overhead of \tool, on the other hand, comes from the GAN model training instead of test sample generation.
As a result, generating various scale numbers of test samples will have no substantial impact on the \tool's total overheads.
The results imply that \texttt{ILFO} is not scalable for testing AdNNs with large datasets, whereas \tool does an excellent job.
We also notice that the \tool's overheads for $L_{2}$ and $L_{inf}$ are different for DN\_SVHN.
Because we use the early stopping method \cite{Yao2007} to train \tool, we can explain such variation in overheads.
In detail, the objective $L_{per}$ differs for $L_2$ and $L_{inf}$. Thus, training process will terminate at different epochs.


\subsection{Effectiveness}
\label{sec:severity}

\subsubsection{Relative Performance Degradation}

\noindent \textbf{Metrics.} To characterize system performance, we choose both hardware-independent and hardware-dependent metrics.
Our hardware independent metric is floating-point operations (FLOPs).
FLOPs are widely used to assess the computational complexity of DNNs~\cite{wu2018blockdrop, wang2018skipnet}. Higher FLOPs indicate higher CPU utilization and lower efficiency performance.
As for hardware-dependent metrics, we focus on latency and energy consumption because these two metrics are essential for real-time applications~\cite{bateni2020neuos, wan2020alert}.
After characterizing system performance with the above metrics, We measure the \textit{increment} in the above performance metrics to reflect the severity of performance degradation.
In particular, we measure the increased percentage of flops~\textit{I-FLOPs}, latency~(\textit{I-Latency}) and energy consumption~(\textit{I-Energy}) as our performance degradation severity evaluation metrics. 

\equref{eq:metric} shows the formal definition of our degradation severity evaluation metrics. In \equref{eq:metric}, $x$ is the original seed input, $\delta$ is the generated perturbation, and $F_{f}(\cdot), \space L_{f}(\cdot), \space E_{f}(\cdot)$ are the functions that measure FLOPs, latency, and energy consumption of AdNN $f(\cdot)$.
A test sample is more effective in triggering performance degradation if it increases more percentage of FLOPs, latency, and energy consumption.
We examine two scenarios for each evaluation metric: the \textit{average} metric value for the whole test dataset and the \textit{maximum} metric value caused for a particular sample.
The first depicts long-term performance degradation, whereas the second depicts performance degradation under the worst-case situation.
We measure the energy consumption using TX2's power monitoring tool~\cite{tx2}. 
\begin{equation}
\label{eq:metric}
\begin{split}
    & I-FLOPs(x) = \frac{F_{f}(x + \delta) - F_{f}(x)}{F_{f}(x)} \times 100 \%\\
    & I-Latency(x) = \frac{L_{f}(x + \delta) - L_{f}(x)}{L_{f}(x)} \times 100 \%\\
    & I-Energy(x) = \frac{E_{f}(x + \delta) - E_{f}(x)}{E_{f}(x)} \times 100 \% \\
\end{split}
\end{equation}

\vspace{-2mm}
\begin{table}[tph!]
\caption{The FLOPs increment of the test samples (\%)}
\label{tab:flops_inc}
\vspace{-3mm}
\resizebox{0.38\textwidth}{!}{
    \begin{tabular}{c|c||rr|rr}
    \toprule
    \multirow{2}[3]{*}{\textbf{Norm}} & \multirow{2}[3]{*}{\textbf{Subject}} & \multicolumn{2}{c|}{\textbf{Mean}} & \multicolumn{2}{c}{\textbf{Max}} \\
\cline{3-6}          &       & \multicolumn{2}{c|}{\textbf{baseline  / ours}} & \multicolumn{2}{c}{\textbf{baseline  /  ours}} \\ \midrule
          & \textbf{SN\_C10} & 6.43  & \textbf{31.14} & 18.43 & \textbf{62.77} \\
          & \textbf{BD\_C10} & \textbf{48.44} & 38.39 & 162.58 & \textbf{188.60} \\
    \textbf{$L_{inf}$} & \textbf{RN\_C100} & 133.67 & \textbf{181.57} & \textbf{498.29} & \textbf{498.99} \\
          & \textbf{DS\_C100} & 116.19 & \textbf{157.66} & 287.98 & \textbf{552.00} \\
          & \textbf{DS\_SVHN} & 115.99 & \textbf{228.32} & \textbf{498.29} & \textbf{498.29} \\
    \hline
          & \textbf{SN\_C10} & 20.34 & \textbf{31.30} & 30.43 & \textbf{82.09} \\
          & \textbf{BD\_C10} & \textbf{48.44} & 38.39 & 162.58 & \textbf{188.60} \\
    \textbf{$L_2$} & \textbf{RN\_C100} & 133.67 & \textbf{182.12} & \textbf{498.99} & \textbf{498.99} \\
          & \textbf{DS\_C100} & 116.19 & \textbf{157.66} & 287.98 & \textbf{552.00} \\
          & \textbf{DS\_SVHN} & 115.99 & \textbf{228.32} & \textbf{498.29} & \textbf{498.29} \\
    \bottomrule
    \end{tabular}%
    \vspace{-5mm}

}
\end{table}


\begin{table*}[htbp!]
\caption{The performance degradation on two hardware platforms (\%)}
\vspace{-3mm}
\label{tab:eng_inc}
\resizebox{0.88\textwidth}{!}{
    \begin{tabular}{c|c||cccc|cccc||cccc|cccc}
    \toprule
    \multirow{4}[6]{*}{\textbf{Device}} & \multirow{4}[6]{*}{\textbf{Subject}} & \multicolumn{8}{c||}{\textbf{L2}}                              & \multicolumn{8}{c}{\textbf{Linf}} \\
\cline{3-18}          &       & \multicolumn{4}{c|}{\textbf{I-Latency}} & \multicolumn{4}{c||}{\textbf{I-Energy}} & \multicolumn{4}{c}{\textbf{I-Latency}} & \multicolumn{4}{c}{\textbf{I-Energy}} \\
\cline{3-18}          &       & \multicolumn{2}{c}{\textbf{Mean}} & \multicolumn{2}{c|}{\textbf{Max}} & \multicolumn{2}{c}{\textbf{Mean}} & \multicolumn{2}{c||}{\textbf{Max}} & \multicolumn{2}{c}{\textbf{Mean}} & \multicolumn{2}{c|}{\textbf{Max}} & \multicolumn{2}{c}{\textbf{Mean}} & \multicolumn{2}{c}{\textbf{Max}} \\
          &       & \multicolumn{2}{c}{\textbf{baseline/ours}} & \multicolumn{2}{c|}{\textbf{baseline/ours}} & \multicolumn{2}{c}{\textbf{baseline/ours}} & \multicolumn{2}{c||}{\textbf{baseline/ours}} & \multicolumn{2}{c}{\textbf{baseline/ours}} & \multicolumn{2}{c|}{\textbf{baseline/ours}} & \multicolumn{2}{c}{\textbf{baseline/ours}} & \multicolumn{2}{c}{\textbf{baseline/ours}} \\
    \midrule
          & \textbf{SN\_C10} & 8.2   & \textbf{25.4} & 20.9  & \textbf{45.7} & 8.3   & \textbf{25.7} & 20.6  & \textbf{44.9} & 5.7   & \textbf{30.9} & 15.1  & \textbf{46.1} & 5.7   & \textbf{31.4} & 15.6  & \textbf{45.8} \\
          & \textbf{BD\_C10} & \textbf{28.7} & 17.5  & \textbf{142.1} & 132.5 & \textbf{28.9} & 17.7  & \textbf{148.2} & 135.3 & 25.4  & \textbf{25.6} & \textbf{143.9} & 135.7 & 25.8  & \textbf{26.1} & 148.2 & \textbf{141.0} \\
    \textbf{CPU} & \textbf{RN\_C100} & \textbf{72.2} & 39.9  & \textbf{1654.4} & 624.1 & \textbf{72.5} & 40.3  & \textbf{1685.1} & 633.7 & 53.6  & \textbf{141.1} & 370.2 & \textbf{1313.1} & 54.1  & \textbf{144.3} & 387.1 & \textbf{1341.1} \\
          & \textbf{DS\_C100} & 61.4  & \textbf{133.8} & 216.2 & \textbf{464.0} & 64.6  & \textbf{142.6} & 217.3 & \textbf{471.8} & 52.0  & \textbf{171.5} & 254.5 & \textbf{483.6} & 54.9  & \textbf{180.1} & 282.0 & \textbf{503.9} \\
          & \textbf{DS\_SVHN} & 29.8  & \textbf{210.1} & 392.3 & \textbf{1496.1} & 30.3  & \textbf{214.6} & 398.4 & \textbf{1467.7} & 70.2  & \textbf{257.2} & 1371.2 & \textbf{1580.8} & 71.7  & \textbf{260.1} & 1372.8 & \textbf{1548.2} \\ \midrule
          & \textbf{SN\_C10} & 4.4   & \textbf{14.3} & 6.8   & \textbf{17.9} & 5.3   & \textbf{15.4} & 8.1   & \textbf{19.7} & 4.4   & \textbf{11.8} & 4.8   & \textbf{15.7} & 5.2   & \textbf{12.4} & 6.1   & \textbf{15.9} \\ 
          & \textbf{BD\_C10} & 9.3   & \textbf{9.8} & \textbf{53.6} & 41.6  & 10.2  & \textbf{11.7} & \textbf{59.0} & 42.4  & 13.9  & \textbf{16.9} & 39.9  & \textbf{41.1} & 15.1  & \textbf{20.2} & 46.4  & \textbf{46.6} \\ 
    \textbf{GPU} & \textbf{RN\_C100} & \textbf{90.6} & 51.0  & \textbf{1968.5} & 923.9 & \textbf{96.9} & 55.4  & \textbf{2446.5} & 1043.4 & 66.9  & \textbf{167.2} & 454.8 & \textbf{1496.8} & 70.6  & \textbf{197.4} & 557.9 & \textbf{1837.4} \\
          & \textbf{DS\_C100} & 56.1  & \textbf{102.7} & 184.9 & \textbf{370.2} & 62.8  & \textbf{116.4} & 194.1 & \textbf{478.8} & 71.7  & \textbf{158.6} & 183.8 & \textbf{384.7} & 80.3  & \textbf{177.9} & 217.1 & \textbf{457.8} \\
          & \textbf{DS\_SVHN} & 11.5  & \textbf{75.9} & 149.7 & \textbf{244.2} & 15.8  & \textbf{92.0} & 172.3 & \textbf{298.3} & 38.7  & \textbf{72.0} & 280.0 & \textbf{308.9} & 47.4  & \textbf{88.0} & 348.3 & \textbf{382.8} \\
    \bottomrule
    \end{tabular}%
        
}
\end{table*}

The hardware-independent experimental results are listed in Table \ref{tab:flops_inc}.
As previously stated, greater I-FLOPs implies that the created test samples demand more FLOPs, which will result in significant system performance reduction.
The table concludes that \tool generates test samples that can cause more severe performance degradation.
Other than that, we have multiple observations. 
First, for four of the five subjects, \tool generates test samples that require more FLOPs, \eg 31.14\%-62.77\% for SN\_C10.
Second, for both $L_2$ and $L_{inf}$ perturbation, the model would require more FLOPs, and the difference between $L_2$ and $L_{inf}$ setting is minimal.
Third, the maximum FLOPs are far greater than the average case for some extreme scenarios, \eg for DS\_SVHN, and DS\_C100.
The hardware-dependent experimental results are listed in Table \ref{tab:eng_inc}. 
Similar to hardware-independent experiments, \tool outperforms \ILFO on 65 out of 80 comparison scenarios. However, for the other 15 comparisons, we explain the results as the following two reasons: (i) the system noise has affected the results because for almost all scenarios \tool has been able to increase more I-FLOPs than \ILFO. (ii) recall in Table \ref{tab:study}, $RN\_C100$ has the the PCCs around 0.64, and the FLOPs increment of $RN\_C100$ for \tool and \texttt{ILFO} is around the same level. Thus,  \tool may cause slightly less latency and energy consumption degradation than \texttt{ILFO}.
However, for $SN\_C10$, although it has low PCCs, \tool can increase much more FLOPs than \texttt{ILFO}, thus, \tool can cause more severe performance degradation.
Based on the results in Table \ref{tab:eng_inc}, we conclude that \tool outperforms baseline in creating inputs that consume higher energy or time.

\begin{figure}[hbtp]
    \centering
    \includegraphics[width=0.42\textwidth]{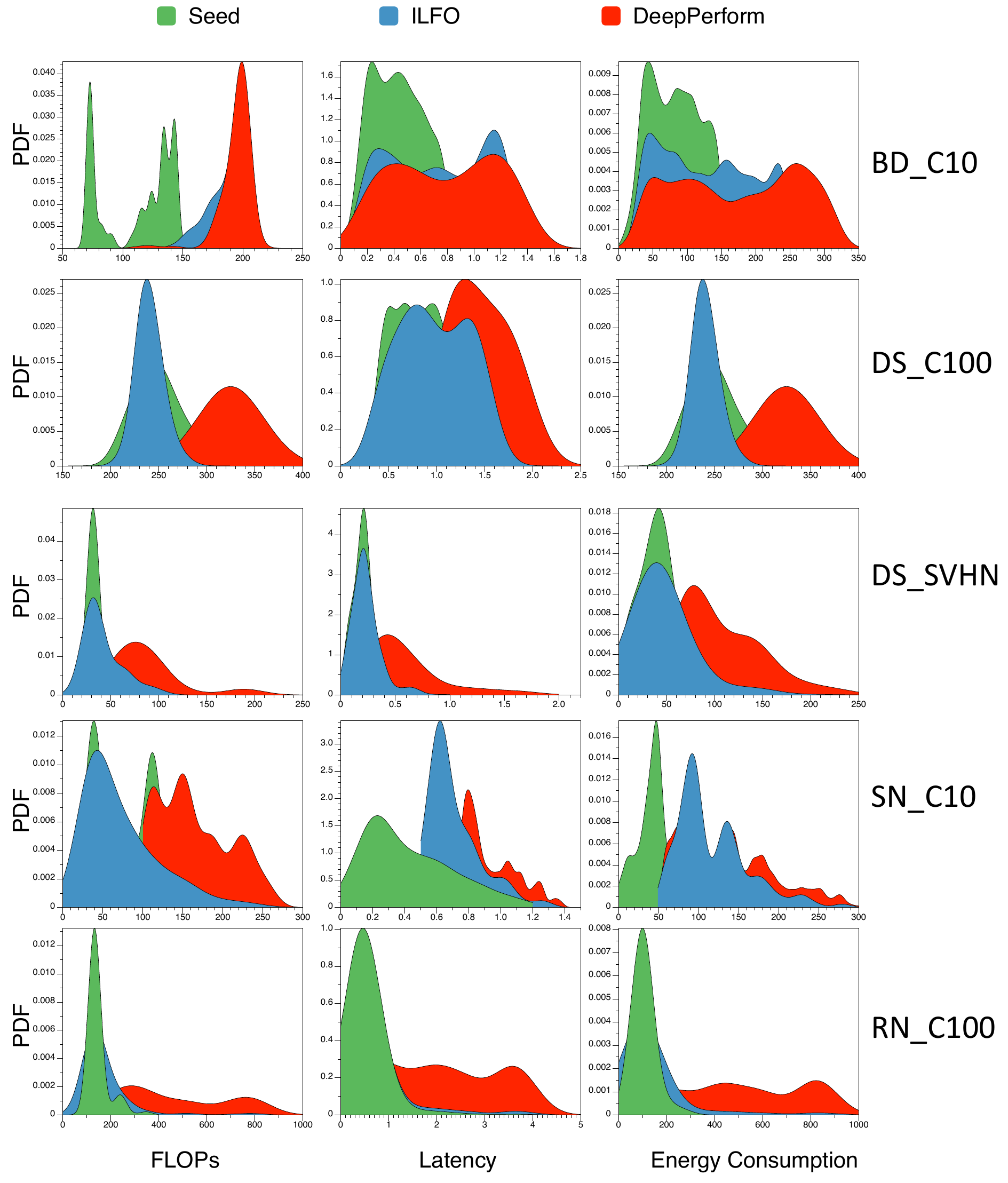}
    \vspace{-3mm}
    \caption{The unnormalized efficiency distribution of seed inputs and the generated inputs}
    \label{fig:distribution}
\end{figure}

\subsubsection{Absolute Performance Degradation}
Besides the relative performance degradation, we also investigate the absolute performance degradation of the generated inputs.
In Figure~\ref{fig:distribution}, we plot the unnormalized efficiency distribution (\ie FLOPs, latency, energy consumption) of both seed and generated inputs to characterize the absolute performance degradation. We specifically depict the probability distribution function (PDF) curve \cite{statistic} of each efficiency metric under discussion.
The unnormalized efficiency distribution is shown in \figref{fig:distribution}, where the green curve is for the seed inputs, and the red curve is for the test inputs from \tool. 
From the results, we observe that \tool is more likely to generate test inputs located at the right end of the x-axis.
Recall that a PDF curve with more points on the right end of the x-axis is more likely to generate theoretical worst-case test inputs.
The results confirm that \tool is more likely to generate test inputs with theoretical worst-case complexities.

\subsubsection{Test Sample Validity}
To measure the validity of the generated test samples, we define \textit{degradation success number} $\eta$ in \equref{eq:validity}, 
\begin{equation}
    \eta = \sum \mathbb I( FLOPs(x + \delta) \ge FLOPs(x)). \qquad\qquad  x \in \mathcal{X}
    \label{eq:validity}
\end{equation}
where $\mathcal{X}$ is the set of randomly selected seed inputs and $\mathbb I( FLOPs(x + \delta) > FLOPs(x))$ indicates whether generated test samples require more computational resources than the seed inputs.
We run \tool and baselines the same experimental time and generate the different number of test samples~($\mathcal{X}$ in \equref{eq:validity}), we then measure $\eta$ in the generated test samples.
For convince, we set the experimental time as the total time of \tool generating 1,000 test samples (same time for \texttt{ILFO}).
From the third column in \tabref{tab:coverage}, we observe that for most experimental settings, \tool achieves a higher degradation success number than \texttt{ILFO}.
Although \texttt{ILFO} is an end-to-end approach, the high overheads of \texttt{ILFO}  disable it to generate enough test samples.


\begin{figure*}[tp]
    \centering
    \includegraphics[width=0.88\textwidth]{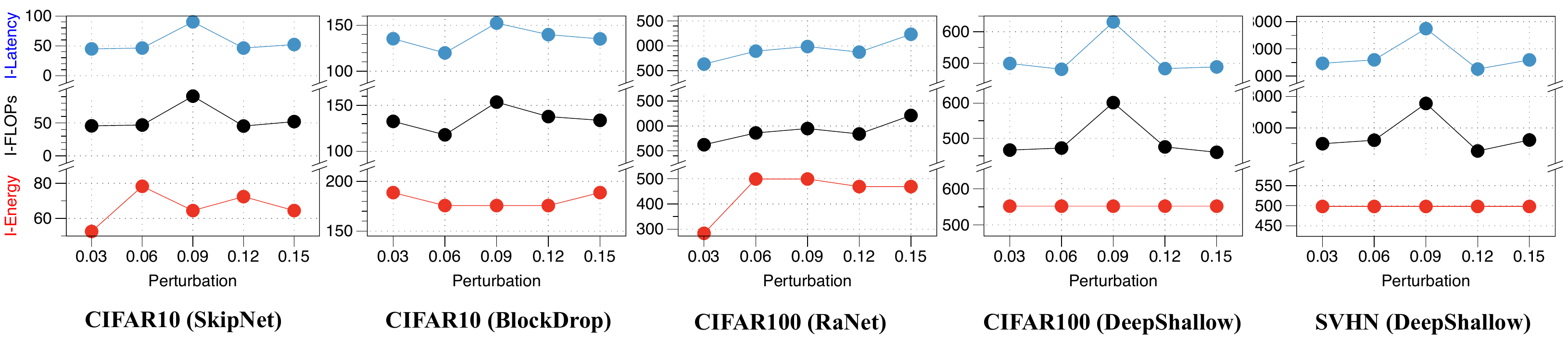}
    \caption{How performance degradation as perturbation constraints change}
    \label{fig:perturbation}
\end{figure*}

\subsection{Coverage}

\begin{table}[htbp]
  \centering
    \vspace{-3mm}
  \caption{Validity and coverage results}
  \vspace{-3mm}
  \resizebox{0.36\textwidth}{!}{
          \begin{tabular}{c|c|cc|cc}
    \toprule
    \multirow{2}[2]{*}{\textbf{Norm}} & \multirow{2}[2]{*}{\textbf{Subject}} & \multicolumn{2}{c|}{\textbf{$\eta$ (\#)}} & \multicolumn{2}{c}{\textbf{$Cov$ (\%)}} \\
          &       & \multicolumn{2}{c|}{\textbf{ours / baseline }} & \multicolumn{2}{c}{\textbf{ ours / baseline}} \\
    \midrule
          & \textbf{SN\_C10} & \textbf{842} & 69    & \textbf{0.74 ± 0.001} & 0.65 ± 0.001 \\
          & \textbf{BD\_C10} & \textbf{630} & 84    & \textbf{0.37 ± 0.001} & 0.37 ± 0.001 \\
    \textbf{Linf} & \textbf{RN\_C100} & \textbf{871} & 133   & \textbf{0.99 ± 0.002} & 0.89 ± 0.030 \\
          & \textbf{DS\_C100} & \textbf{646} & 69    & \textbf{1.00 ± 0.000} & 0.83 ± 0.016 \\
          & \textbf{DS\_SVHN} & \textbf{916} & 220   & \textbf{1.00 ± 0.000} & 0.92 ± 0.033 \\
    \midrule
          & \textbf{SN\_C10} & \textbf{993} & 81    & 0.84 ± 0.001 & \textbf{0.85 ± 0.001} \\
          & \textbf{BD\_C10} & \textbf{732} & 79    & \textbf{0.41 ± 0.001} & 0.40 ± 0.001 \\
    \textbf{L2} & \textbf{RN\_C100} & \textbf{924} & 229   & 0.94 ± 0.007 & \textbf{0.95 ± 0.013} \\
          & \textbf{DS\_C100} & \textbf{734} & 181   & \textbf{1.00 ± 0.000} & \textbf{1.00 ± 0.000} \\
          & \textbf{DS\_SVHN} & \textbf{924} & 518   & \textbf{0.98 ± 0.025} & 0.73 ± 0.034 \\
    \bottomrule
    \end{tabular}%
}
  \label{tab:coverage}%
\end{table}%

In this section, we investigate the comprehensiveness of the generated test inputs. In particular, we follow existing work \cite{pei2017deepxplore, zhang2018deeproad} and investigate the diversity of the AdNN behaviors explored by the test inputs generated by \tool.
Because AdNNs' behavior relies on the computation of intermediate states \cite{pei2017deepxplore,ma2018deepgauge}, we analyze how many intermediate states are covered by the test suite.
\begin{equation}
     \quad   Cov(\mathcal{X}) = \frac{\sum_{x \in \mathcal{X}}\sum_{i=1}^N \mathbb I( B_i(x) > \tau_i))}{N} 
    \label{eq:coverage}
\end{equation}

\noindent To measure the coverage of AdNNs' intermediate states, we follow existing work~\cite{pei2017deepxplore} and define decision \textit{block coverage} ($Cov(\mathcal{X})$ in \equref{eq:coverage}),
where $N$ is the total number blocks, $\mathbb I(\cdot)$ is the indicator function, and
$(B_i(x) > \tau_i))$ represents whether $i^{th}$ block is activated by input $x$~(the definition of $B_i$ and $\tau_i$ are the same with \equref{eq:new2} and \equref{eq:new3}).
Because AdNNs activate different blocks for decision making, then a higher block coverage indicates the test samples cover more decision behaviors. 
For each subject, we randomly select 100 seed samples from the test dataset as seed inputs. We then feed the same seed inputs into \tool and \texttt{ILFO} to generate test samples. 
Finally, we feed the generated test samples to AdNNs and measure block coverage.
We repeat the process ten times and record the average coverage and the variance.
The results are shown in \tabref{tab:coverage} last two columns. 
We observe that the test samples generated by \tool achieve higher coverage for almost all subjects.


\subsection{Sensitivity}
\label{sec:sensitive}

In this section, we conduct two experiments to show that \tool can generate effective test samples under different settings.

\noindent\textbf{Configuration Sensitivity.} As discussed in \secref{sec:background}, AdNNs require configuring the threshold $\tau_i$ to set the accuracy-performance tradeoff mode. In this section, we evaluate whether the test samples generated from \tool could degrade the AdNNs' performance under different modes.
Specifically, we set the threshold $\tau_i$ in \equref{eq:new2} and \equref{eq:new3} as $0.3, 0.4, 0.5, 0.6, 0.7$ and measure the maximum FLOPs increments. 
Notice that we train \tool with $\tau_i=0.5$ and test the performance degradation with different $\tau_i$.
The maximum FLOPs increment ratio under different system configurations are listed in \tabref{tab:configure}.
For all experimental settings, the maximum FLOPs increment ratio keeps a stable value~(\eg 79.17-82.91, 175.59-250.00).
The results imply that the test samples generated by \tool can increase the computational complexity under different configurations, and the maximum FLOPs increment ratio is stable as the configuration changes.

\begin{table}[htp]
  \centering
  \caption{Increment under different thresholds}
  \resizebox{0.38\textwidth}{!}{
         \begin{tabular}{c|c|rrrrr}
    \toprule
    \multirow{2}[2]{*}{\textbf{Norm}} & \multirow{2}[2]{*}{\textbf{Subject}} & \multicolumn{5}{c}{\textbf{Threshold}} \\
          &       & \textbf{0.3} & \textbf{0.4} & \textbf{0.5} & \textbf{0.6} & \textbf{0.7} \\
    \midrule
          & \textbf{SN\_C10} & 79.17  & 82.91  & 82.91  & 75.00  & 70.00  \\
          & \textbf{BD\_C10} & 250.00  & 250.00  & 175.59  & 175.59  & 175.59  \\
    \textbf{L2} & \textbf{RN\_C100} & 500.00  & 498.99  & 498.99  & 200.00  & 200.00  \\
          & \textbf{DS\_C100} & 600.00  & 600.00  & 552.00  & 400.00  & 200.00  \\
          & \textbf{DS\_SVHN} & 498.29  & 498.29  & 498.29  & 498.29  & 400.00  \\
    \midrule
          & \textbf{SN\_C10} & 66.67  & 78.26  & 82.91  & 66.67  & 73.91  \\
          & \textbf{BD\_C10} & 233.33  & 175.59  & 175.59  & 233.33  & 233.33  \\
    \textbf{Linf} & \textbf{RN\_C100} & 498.99  & 498.99  & 498.99  & 498.99  & 498.99  \\
          & \textbf{DS\_C100} & 552.00  & 552.00  & 552.00  & 400.00  & 300.00  \\
          & \textbf{DS\_SVHN} & 498.29  & 498.29  & 498.29  & 498.29  & 400.00  \\
    \bottomrule
    \end{tabular}%
}
  \label{tab:configure}%

\end{table}%


\begin{table}[hbp]
  \centering
  \caption{Performance degradation on different hardware}

  \resizebox{0.4\textwidth}{!}{
        \begin{tabular}{c|c|cc|cc}
    \toprule
    \multirow{2}[2]{*}{\textbf{Norm}} & \multirow{2}[2]{*}{\textbf{Subject}} & \multicolumn{2}{c|}{\textbf{Intel Xeon E5-2660 v3 CPU}} & \multicolumn{2}{c}{\textbf{Nvidia 1080 Ti}} \\
          &       & \multicolumn{2}{c|}{\textbf{ I-Latency / I-Energy }} & \multicolumn{2}{c}{\textbf{ I-Latency / I-Energy }} \\
    \midrule
          & \textbf{SN\_C10} & 36.95  & 36.20  & 24.94  & 50.77  \\
          & \textbf{BD\_C10} & 76.69  & 79.24  & 64.10  & 63.55  \\
    \textbf{L2} & \textbf{RN\_C100} & 1019.25  & 1173.21  & 938.21  & 856.46  \\
          & \textbf{DS\_C100} & 567.10  & 609.73  & 414.38  & 338.51  \\
          & \textbf{DS\_SVHN} & 236.12  & 246.70  & 311.01  & 282.09  \\
    \midrule
          & \textbf{SN\_C10} & 29.38  & 28.28  & 24.95  & 11.94  \\
          & \textbf{BD\_C10} & 70.67  & 74.09  & 49.82  & 52.70  \\
    \textbf{Linf} & \textbf{RN\_C100} & 319.72  & 355.29  & 679.79  & 652.98  \\
          & \textbf{DS\_C100} & 463.91  & 496.84  & 439.53  & 464.65  \\
          & \textbf{DS\_SVHN} & 232.88  & 244.91  & 263.49  & 141.56  \\
    \bottomrule
    \end{tabular}%
    }
  \label{tab:hardware}%
\end{table}%

\noindent\textbf{Hardware Sensitivity.} We next evaluate the effectiveness of our approach on different hardware platforms. In particular, we select Intel Xeon E5-2660 V3 CPU and Nvidia 1080 Ti as our experimental hardware platforms and measure the maximum performance degradation ratio on those selected platforms. The test samples generated by \tool, as shown in \tabref{tab:hardware},  cause severe and stable runtime performance degradation on different hardware platforms. As a result, we conclude that \tool is not sensitive to hardware platforms.

\subsection{Quality}
\label{sec:perturbation}

\begin{table}[hbp]
  \centering
  \caption{The perturbation size of the generated test inputs}
  \vspace{-3mm}
  \resizebox{0.4\textwidth}{!}{
    \begin{tabular}{c|ccccc}
    \toprule
    \textbf{Norm} & \multicolumn{1}{c}{\textbf{SN\_C10}} & \multicolumn{1}{c}{\textbf{BD\_C10}} & \multicolumn{1}{c}{\textbf{RN\_C100}} & \multicolumn{1}{c}{\textbf{DS\_C100}} & \multicolumn{1}{c}{\textbf{DS\_SVHN}} \\
    \midrule
    \textbf{L2} & 9.48  & 9.47  & 9.50 & 9.48  & 9.62  \\
    \textbf{Linf} & 0.03  & 0.03  & 0.03  & 0.03  & 0.03  \\
    \bottomrule
    \end{tabular}%
}
  \label{tab:per}%
\end{table}%

\begin{figure}[htb!]
    \centering
    \includegraphics[width=0.48\textwidth]{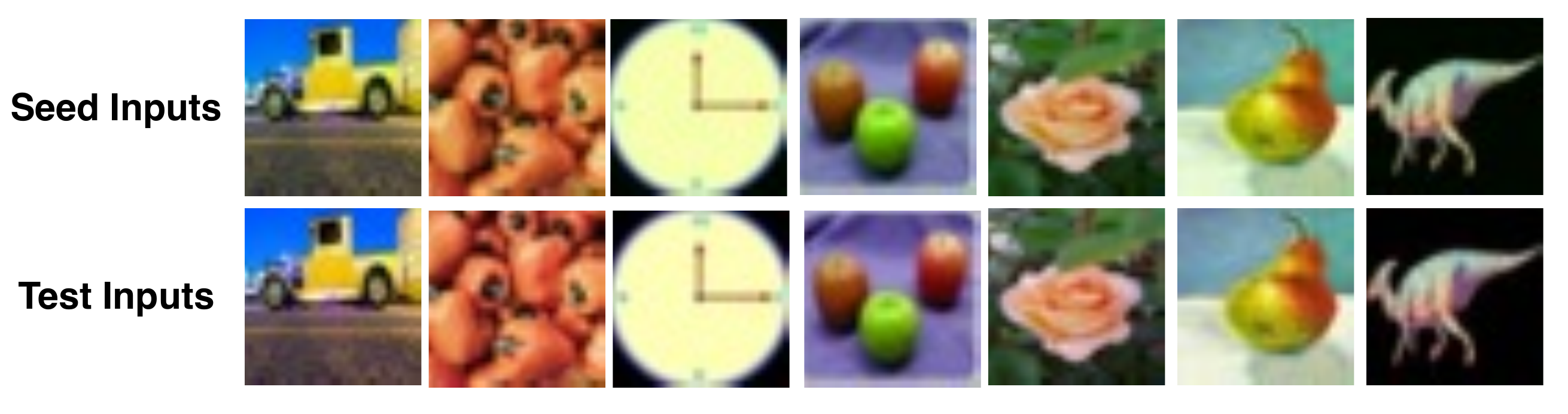}
    \vspace{-3mm}
    \caption{Testing inputs generated by \tool}
    \label{fig:case}
\end{figure}

\noindent We first conduct quantitative evaluations to evaluate the similarity between the generated and seed inputs. In particular, we follow existing work \cite{carlini2017towards} and compute the perturbation magnitude. The perturbation magnitude are listed in \tabref{tab:per}. Recall that we follow existing work \cite{carlini2017towards, MirazILFO} and set the perturbation constraints $\epsilon$ as $10$ and $0.03$ for $L_2$ and $L_{inf}$ norm (\secref{sec:setup}). From the results in \tabref{tab:per}, we conclude that generated test samples can satisfy the semantic-equivalent constraints in \equref{eq:define}.
Moreover, we conduct a qualitative evaluation. In particular, we randomly select seven images from the generated images for RA\_C100 and visualize them in  \figref{fig:case}~(more results are available on our website), where the first row is the randomly selected seed inputs, and the second row is the corresponding generated inputs. The visualization results show that the test inputs generated by \tool are semantic-equivalent to the seed inputs.
Furthermore, we investigate the relationship between different semantic-equivalent constraints and performance degradation.
We first change the perturbation magnitude constraints (\ie $\eta$ in \equref{eq:define}) and train different models~(experiment results for $L_{2}$ norm could be found on our websites).
After that, we measure the severity of AdNN performance degradation under various settings.
\figref{fig:perturbation} shows the results. We observe that although the relationship between performance degradation ratio and perturbation magnitude is not purely linear, there is a trend that the overhead increases with the increase of perturbation magnitude.


\begin{table}[tbp]
  \centering
  \caption{Efficiency and accuracy of AdNN model}
  \vspace{-3mm}
  \resizebox{0.46\textwidth}{!}{
    \begin{tabular}{cl|ccccc}
    \toprule
    \multicolumn{2}{c|}{\textbf{Metric}} & \textbf{SN\_C10} & \textbf{BD\_C10} & \textbf{RN\_C100} & \textbf{DS\_C100} & \textbf{DS\_SVHN} \\
    \midrule
    \multirow{2}[1]{*}{\textbf{I-FLOPs}} & \textbf{before} & 31.30  & 38.39  & 182.12  & 157.66  & 228.32  \\
          & \textbf{after} & 8.07  & 15.26  & 35.37  & 28.54  & 38.65  \\
    \multirow{2}[1]{*}{\textbf{Acc}} & \textbf{before} & 92.34  & 91.35  & 65.43  & 58.78  & 94.54  \\
          & \textbf{after} & 13.67  & 10.56  & 6.67  & 7.67  & 18.78  \\
    \bottomrule
    \end{tabular}%
    }
  \label{tab:retrain}%
\end{table}%

\section{Application}
\label{sec:app}

This section investigates if developers can mitigate the performance degradation bugs using the existing methods for mitigating DNN correctness bugs (\ie adversarial examples).
We focus on two of the most widely employed approaches: offline adversarial training \cite{goodfellow2014explaining}, and online input validation \cite{wang2020dissector}.
Surprisingly, we discover that not all of the two approaches can address performance faults in the same manner they are used to repair correctness bugs.

\subsection{Adversarial Training}
\noindent\textbf{Setup.} We follow existing work \cite{goodfellow2014explaining} and feed the generated test samples and the original model training data to retrain each AdNN.
The retraining objective can be formulated as 
\begin{equation}
    \mathcal{L}_{retrain} = \underbrace{\ell(g_{f}(x'), \; g_{f}(x))}_{\mathcal{L}_1} + \underbrace{\beta \left\{ \ell(f(x), y) + \ell(f(x'), y) \right\}}_{\mathcal{L}_2}
    \label{eq:retrain}
\end{equation}
\noindent where $x$ is one seed input in the training dataset, $x' = \mathcal{G}(x) + x$ is the generated test input, $f(\cdot)$ is the AdNNs, and  $g_{f}(\cdot)$ measures the AdNNs computational FLOPs.
Our retraining objective can be interpreted as forcing the buggy test inputs $x'$ to consume the same FLOPs as the seed one (\ie $\mathcal{L}_1$), while producing the correct results (\ie $\mathcal{L}_2$).
For each AdNN model under test, we retrain it to minimize the objective in \equref{eq:retrain}.
After retraining, we test each AdNNs accuracy and efficiency on the hold-out test dataset.

\noindent\textbf{Results.}
\tabref{tab:retrain} shows the results after model retraining.
The left two columns show the performance degradation before and after model retraining, while the right two columns show the model accuracy before and after model retraining.
The findings show that following model training, the I-FLOPs fall; keep in mind that a higher I-FLOPs signifies a more severe performance degradation. Thus, the decrease in I-FLOPs implies that model retraining can help overcome performance degradation. However, based on the data in the right two columns, we observe that such retraining, different from accuracy-based retraining, may harm model accuracy.

\begin{table}[tbp]
  \centering
  \caption{Performance of SVM detector }
  \vspace{-3mm}
  \resizebox{0.4\textwidth}{!}{
    \begin{tabular}{c|cc|cc|cc}
    \toprule
    \multirow{2}[2]{*}{\textbf{Subject}} & \multicolumn{2}{c|}{\textbf{AUC}} & \multicolumn{2}{c|}{\textbf{Extra Latency (s)}} & \multicolumn{2}{c}{\textbf{Extra Energy (j)}} \\
          & \textbf{L2} & \textbf{Linf} & \textbf{L2} & \textbf{Linf} & \textbf{L2} & \textbf{Linf} \\
    \midrule
    \textbf{SN\_C10} & 0.9997  & 0.9637  & 0.0168  & 0.0167  & 1.8690  & 1.8740  \\
    \textbf{BD\_C10} & 0.9967  & 0.9222  & 0.0001  & 0.0002  & 0.0108  & 0.0197  \\
    \textbf{RN\_C100} & 1.0000  & 0.9465  & 0.0031  & 0.0042  & 0.3263  & 0.4658  \\
    \textbf{DS\_C100} & 0.5860  & 0.3773  & 0.0167  & 0.0212  & 1.8578  & 2.4408  \\
    \textbf{DS\_SVHN} & 1.0000  & 1.0000  & 0.0098  & 0.0210  & 1.1030  & 2.3959  \\
    \bottomrule
    \end{tabular}%
    }
  \label{tab:svm}%
    \vspace{-5mm}
\end{table}%

\subsection{Input Validation}

Input validation \cite{wang2020dissector} is a runtime approach that filters out abnormal inputs before AdNNs cast computational resources on such abnormal inputs.
This approach is appropriate for situations where the sensors (\eg camera) and the decision system (\eg AdNN) work at separate frequencies.
Such different frequency working mode is very common in robotics systems \cite{luo2019grouped,feichtenhofer2019slowfast,zolfaghari2018eco}, where the AdNN system will randomly select one image from continuous frames from sensors since continuous frames contain highly redundant information.
Our intuition is to filter out those abnormal inputs at the early computational stage, the same as previous work \cite{wang2020dissector}.

\noindent\textbf{Design of Input Filtering Detector.} 
Our idea is that although seed inputs and the generated test inputs look similar, the latent representations of these two category inputs are quite different~\cite{wang2020dissector}. 
Thus, we extract the hidden representation of a given input by running the first convolutional layer of the AdNNs. First, we feed both benign and \tool generated test inputs to specific AdNN.
We use the outputs of the first convolutional layer as input to train a linear SVM to classify benign inputs and inputs that require huge computation. If any resource consuming adversarial input is detected, the inference is stopped.
The computational complexity of the SVM detector is significantly less than AdNNs. Thus the detector will not induce significant computational resources consumption. 

\noindent\textbf{Setup.} For each experimental subject, we randomly choose 1,000 seed samples from the training dataset, and apply \tool to generate 1,000 test samples. We use these 2,000 inputs to train our detector. 
To evaluate the performance of our detector, we first randomly select 1,000 inputs from the test dataset and apply \tool to generate 1000 test samples.
After that, we run the trained detector on such 2,000 inputs and measure detectors' AUC score, extra computation overheads, and energy consumption.

\noindent\textbf{Results.}
\tabref{tab:svm} shows that the trained SVM detector can successfully detect the test samples that require substantial computational resources.
Specifically for $L_2$ norm perturbation, all the AUC scores are higher than 0.99. The results indicate that the proposed detector identifies $L_2$ test samples better.
The last four columns show the extra computational resources consumption of the detector. We observe that the detector does not consume many additional computational resources from the results.

\section{Threats To Validity}
Our selection of five experimental subjects might be the \textit{external threat} that threaten the generability of our conclusions.
We alleviate this threat by the following efforts.
(1) We ensure that the datasets are widely used in both academia and industry research. 
(2) All evaluated models are state-of-the-art DNN models (published in top-tier conferences after 2017). 
(3) Our subjects are diverse in terms of a varied set of topics: all of our evaluated datasets and models differ from each other in terms of different input domains (\eg digit, general object recognition), the number of classes (from 10 to 100), the size of the training dataset (from 50,000 to 73,257), the model adaptive mechanism.
Our \textit{internal threat} mainly comes from the realism of the generated inputs.
We alleviate this threat by demonstrating the relationship of our work with existing work.
Existing work \cite{zhang2018deeproad,zhou2020deepbillboard, kurakin2018adversarial} demonstrates that correctness-based test inputs exist in the physical world. Because we formulate our problem(\ie the constraint in \equref{eq:define}) the same as the previous correctness-based work \cite{zhou2020deepbillboard, madry2017towards}, we conclude our generated test samples are real and exist in the physical world.

\section{Related Works}

\noindent\textbf{Adversarial Examples \& DNN Testing.} Adversarial Examples have been used evaluate the robustness of DNNs.
These examples are fed to DNNs to change the prediction of the model. 
Szegedy \textit{et al.} \cite{szegedy2013intriguing} and Goodfellow \textit{et al.} \cite{goodfellow2014explaining} propose adversarial attacks on DNNs. Karmon \textit{et al.} Adversarial attacks have been extended to various fields like natural language and speech processing \cite{carlini2016hidden,jia2017adversarial}, and graph models \cite{zugner2018adversarial,bojchevski2019adversarial}. Although, all these attacks focus on changing the prediction and do not concentrate on performance testing.
Several testing methods have been proposed to test DNNs~\cite{zhang2018deeproad, zhou2020deepbillboard, chen2022nicgslowdown, chen2021transslowdown}.

\noindent\textbf{Performance Testing.} Runtime performance is a critical property  of software, and a branch of work has been proposed to test software performance.
For example, Netperf~\cite{netp} and IOZone~\cite{iozone} evaluate the performance of different virtual machine technologies. 
\texttt{WISE} \cite{burnim2009wise} proposes a method to generate test samples to trigger worst-case complexity.
\texttt{SlowFuzz} \cite{petsios2017slowfuzz} proposes a fuzzing framework to detect algorithmic complexity vulnerabilities.
\texttt{PerfFuzz} \cite{lemieux2018perffuzz}
generates inputs that trigger pathological behavior across program locations.

\section{Conclusion}

In this paper, we propose \tool, a performance testing framework for DNNs. Specifically, \tool trains a GAN to learn  and  approximate  the distribution of  the  samples  that  require  more  computational units. Through our evaluation, we have shown that \tool is able to find IDPB in AdNNs more effectively and efficiently than baseline techniques.

\begin{acks}
This work was partially supported by Siemens Fellowship and NSF grant CCF-2146443.
\end{acks}

\bibliographystyle{ACM-Reference-Format}
\bibliography{egbib}

\end{document}